\documentclass{lmcs} 
\pdfoutput=1
\usepackage[utf8]{inputenc}

\usepackage{lastpage}
\lmcsdoi{21}{3}{31}
\lmcsheading{}{\pageref{LastPage}}{}{}%
{Apr.~04,~2022}{Sep.~24,~2025}{}


\usepackage{hyperref}
\usepackage{amsmath}
\usepackage{amsthm}
\usepackage{graphicx}
\usepackage{xcolor}

\usepackage{algorithm}
\usepackage{algorithmic}

\begin{document}

\title[Modeling Software Behavior by Learning Probabilistic Automata]{FlexFringe: Modeling Software Behavior by Learning Probabilistic Automata}

\author[S.Verwer]{Sicco Verwer\lmcsorcid{0000-0002-3682-0962}}
\address{Department of Software Technology, Delft University of Technology, the Netherlands}
\email{s.e.verwer@tudelft.nl}

\author[C.Hammerschmidt]{Christian Hammerschmidt}

\begin{abstract}
We present the efficient implementations of probabilistic deterministic finite automaton learning methods available in FlexFringe. These are well-known strategies for state merging, including several modifications to improve their performance in practice. We show experimentally that these algorithms obtain competitive results and significant improvements over a default implementation. We also demonstrate how to use FlexFringe to learn interpretable models from software logs and use these for anomaly detection. Although less interpretable, we show that learning smaller, more convoluted models improves the performance of FlexFringe on anomaly detection, making it competitive with an existing solution based on neural nets.
\end{abstract}

\maketitle

\section{Introduction}

Automata (state machines) are key models for the design and analysis of software systems~\cite{lee1996principles}. Producing and maintaining such models is costly and error-prone, so methods have been developed to learn their structure automatically from trace data~\cite{cook1998discovering}. Learned automata have been used to analyze different types of complex software systems such as web-services~\cite{bertolino2009automatic,ingham2007learning}, X11 programs~\cite{ammons2002mining}, communication protocols~\cite{comparetti2009prospex,antunes2011reverse,fiteruau2017model, fiterau2020analysis}, Java programs~\cite{cho2011mace}, and malicious software~\cite{cui2007discoverer}. These studies highlight a great benefit that automata provide over more traditional machine learning models, which is that the behavior of many different software systems can be succinctly modeled using different kinds of automata. Through visualization, these models give unparalleled insight into the inner workings of software systems and have been used to discover and locate different types of software errors.

FlexFringe~\cite{verwer2017flexfringe}, which originated from the DFASAT~\cite{heule2010exact} algorithm, is a framework for learning different kinds of automata using the red-blue state merging framework~\cite{lang1998results}. Learning automata from traces can be seen as a grammatical inference~\cite{higuera2010book} problem where traces are modeled as the words of a language, and the goal is to find a model for this language, e.g., a (probabilistic) deterministic finite state automaton (P)DFA~\cite{hopcroft2001introduction}. Although the problem of learning a (P)DFA is NP-hard~\cite{gold1978complexity} and hard to approximate~\cite{pitt1993minimum}, state merging is a well-known and effective heuristic method for solving this problem~\cite{lang1998results}. 

\begin{figure}
    \centering
    \includegraphics[width=\textwidth]{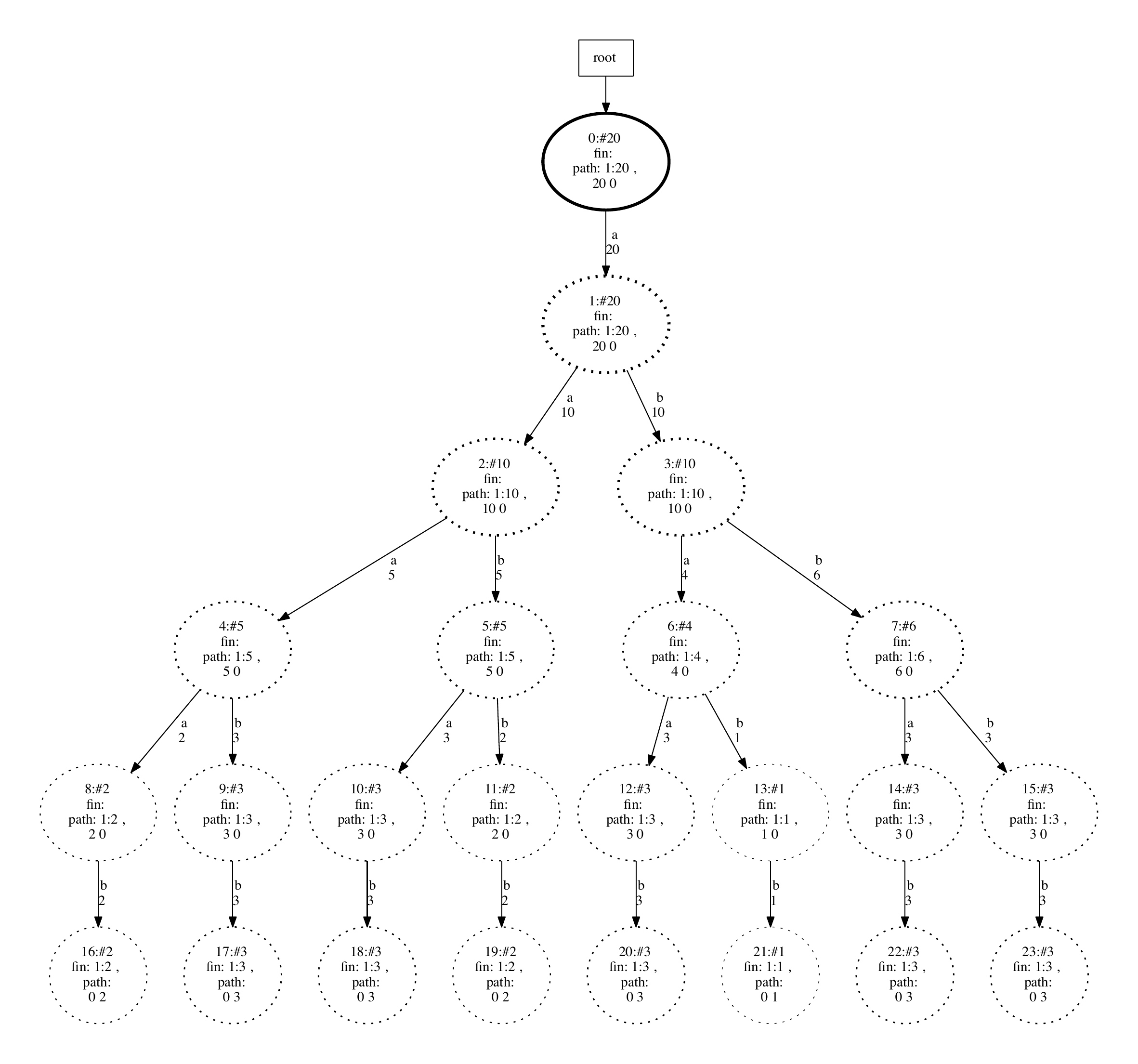}
    \caption{A prefix tree printed by FlexFringe and displayed using Graphviz dot. Each state contains a state number, occurrence counters for the total number of traces (\#), as well as how many of them end (fin counts in format type:count) or pass through (path counts in format type:count) for each trace type (in this case positive - type 1). The last two numbers in each state are the sums of path and fin counts over all types. One can use these to infer what traces occurred in the training data, e.g., a-a-a-b-b occurred 3 times, following transitions from state 0-1-2-4-9 and ending in state 17. Transitions are labeled by symbols and occurrence counts. The initial state has a solid edge, indicating it has already been learned as an automaton state. The dotted states can still be merged with both solid and dotted states. Note that this is the default print setting. One can modify what is printed using command-line parameters or by defining a new printing function.}
    \label{fig:learning_example}
\end{figure}

\begin{figure}
    \centering
    \includegraphics[width=0.6\textwidth]{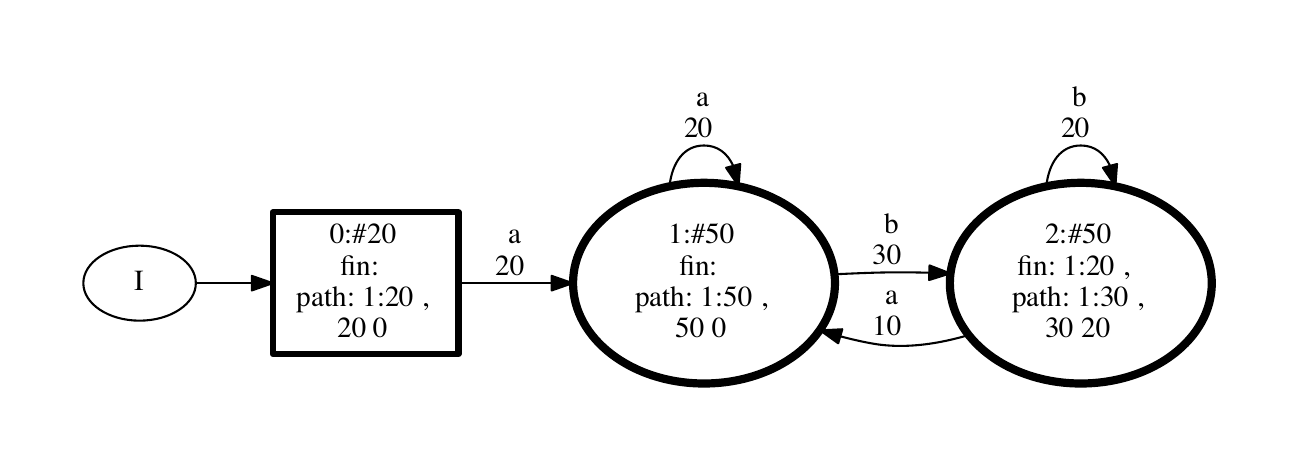}
    \includegraphics[width=0.5\textwidth]{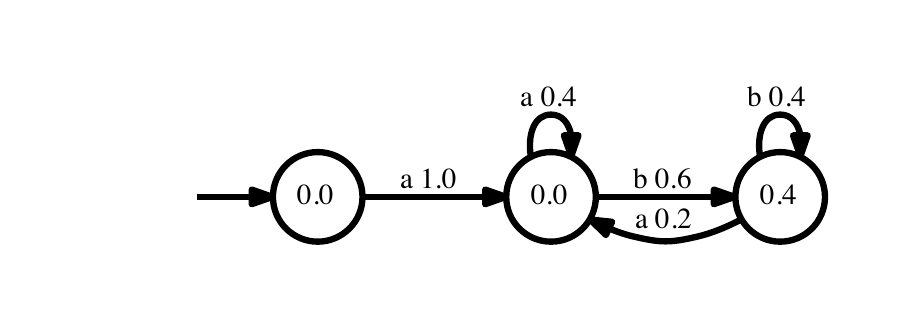}
    \caption{An automaton model printed after running FlexFringe (top). It contains the same type of counts as the prefix tree. To obtain a PDFA from these counts, one needs to normalize them to obtain transition and final probabilities (bottom). Traces only end in the third state, making it the only possible ending state. The learned PDFA, therefore, correctly represents the set of traces starting with ``a'' and ending in ``b'', i.e., $a(a|b)^*b$. A learned PDFA can assign probabilities to new sequences by following transitions and multiplying their probabilities. It can also be used for anomaly detection, for instance, by checking whether a new trace ends in a state with a final probability of 0.}
    \label{fig:learning_example_result}
\end{figure}

State-merging starts with a large tree-shaped model called the prefix tree, which directly encodes the input traces. It then iteratively combines states by testing the similarity of their future behaviors using a Markov property~\cite{norris1998markov} or a Myhill-Nerode congruence~\cite{hopcroft2001introduction}. This process continues until no similar states can be found. The result is a small model displaying the system states and transition structure hidden in the data. Figure~\ref{fig:learning_example} shows the prefix tree for a small example data set consisting of 20 sequences starting with an ``a'' event and ending in a ``b'' event. Running FlexFringe results in the PDFA shown in Figure~\ref{fig:learning_example_result}.

Interestingly, we see that a DFA acceptor for a language (or classifier) can be learned from only positive (also called unlabeled or unsupervised) data. The state-merging algorithm uses statistical tests to discover the states and transition structure of a PDFA. The labels only provide meaning to these states, converting it into a DFA. In practice, it is often easier to learn probabilistic automata since we only observe and store traces from software executions that actually occur, i.e., positive data. Often, we do not have access to negative (non-occurring) data. The PDFA learning methods implemented in FlexFringe are therefore applicable to many kinds of software systems, see, e.g.,~\cite{nadeem2021alert, lin2020safety, cao2022learning, yang2019improving}

A fundamental design principle of FlexFringe is that this state merging approach can be used to learn models for a wide range of systems, including Mealy and I/O machines~\cite{shahbaz2009inferring,aarts2010learning}, probabilistic deterministic automata~\cite{vidal2005probabilistic, verwer2014pautomac}, timed/extended automata~\cite{verwer2010efficient, niggemann2012learning, niggemann2012learning, schmidt2014online, walkinshaw2016inferring}, and regression automata~\cite{lin2016short}. All that needs to be modified is the similarity test, implemented as an evaluation function that tests for merge consistency and computes a score function to decide between possible consistent merges. 

The different PDFA learning algorithms in FlexFringe are implemented by only adding a single file containing these functions, including Alergia~\cite{carrasco1994learning}, MDI~\cite{thollard2000probabilistic}, Likelihood-ratio~\cite{verwer2010likelihood}, and AIC minimization~\cite{verwer2010efficient}.

Our main goal is to get a wider audience interested in the efficient state-merging algorithms and their implementation in FlexFringe. Therefore, we highlight the flexible and interpretable nature of FlexFringe and its efficiency and practical performance. In previous work~\cite{verwer2017flexfringe}, we presented FlexFringe and showed how it can be used to learn non-probabilistic automata for bug discovery. In this work, we dive deeper into the probabilistic automata learning algorithms and show:

\begin{itemize}
    \item details of the implementation of the state merging algorithm in FlexFringe, such as the used union/find data structure,
    \item improvements in basic state-merging routines that increase their efficiency and performance,
    \item a comparison of the PDFA learning methods in FlexFringe on the PAutomaC competition data~\cite{verwer2014pautomac},
    \item how to use learned PDFA models to detect behavioral anomalies in a software system (HDFS~\cite{xu2009detecting}) - outperforming an existing solution based on neural nets~\cite{du2017deeplog},
    \item and how to use learned PDFA models to analyze the discovered anomalous patterns.
\end{itemize}

This paper is organized as follows. We start with an overview of PDFAs and the state-merging algorithm in Sections~\ref{sec:PDFAs} and~\ref{sec:FlexFringe}, including a description of the efficient data structures used in FlexFringe. We then describe the implemented PDFA evaluation functions in Section~\ref{sec:functions} and the developed improvements in Section~\ref{sec:improvements}. The results obtained using the different evaluation functions on the PAutomaC competition data sets are discussed in Section~\ref{sec:results}. For software log data, we show the performance on the HDFS data from both insight and performance perspectives in Section~\ref{sec:results_hdfs}. We provide an overview of closely related algorithms and tools in Section~\ref{sec:related}, and end with some concluding remarks in Section~\ref{sec:conclusions}. 

\section{Probabilistic Automata}\label{sec:PDFAs}

Automata or state machines are models for sequential behavior. Like hidden Markov models~\cite{rabiner1986introduction}, they are models with hidden/latent state information. Given a string (trace) of observed symbols (events) $a_1, \ldots, a_n$, this means that the corresponding system states $q_0, q_1, \ldots, q_n$ are unknown/unobserved. A probabilistic automaton models a probability distribution over such state sequences and, hence, over sequences of symbols (event traces). Below, we give definitions of probabilistic automaton models and explain how they are used to assign probabilities. We start with non-deterministic probabilistic automata and describe both HMMs and deterministic probabilistic automata as restrictions thereof.

\begin{defi} \sloppy 
A probabilistic finite state automaton (PFA) is a tuple $\mathcal{A} = \{\Sigma, Q, I, T, F\}$, where $\Sigma$ is a finite alphabet, $Q$ is a finite set of states, $I : Q \to \left[0,1\right]$ is the initial probability function, $T : Q \times \Sigma \times Q \to \left[0,1\right]$ is the transition probability function, and $F : Q \to \left[0,1\right]$ is the final probability function. The probability functions are such that $\sum_{q \in Q} I(q) = 1$, and for all $q \in Q : F(q) + \sum_{a \in \Sigma, q' \in Q} T(q,a,q') = 1$.
\end{defi} \fussy

Given a sequence of observed symbols $s = a_1, \ldots, a_n$, a PFA $\mathcal{A}$ assigns/computes probabilities by summing up the probabilities of all possible state paths given $s$, where the probability of one state path is the product of initial, transition, and final probabilities: 
\[
\mathcal{A}(s) = \sum_{q_0 \in Q} I(q_0) \left( \sum_{q_1 \in Q} T(q_0,a_1, q_1) \left( \sum_{q_2 \in Q} T(q_1,a_2, q_2) \left( \ldots \sum_{q_n \in Q} T(q_{n-1}, a_n, q_n) F(q_n) \right) \right) \right)
\]

Intuitively, such a path starts in any of the start states $q_0 : I(q_0) > 0$, triggers any of the transitions $T(q_{i-1}, a_i, q_i)$ for $1 \leq I \leq n$, reaches $q_0 \ldots, q_n$, and ends in the final state $q_n : F(q_n) > 0$. An exponential number of such paths exist. Still, it can be computed in $O(n|Q|^2)$ run-time using a variant of the forward-backward algorithm for computing probabilities in an HMM~\cite{rabiner1986introduction}. In fact, an HMM can be viewed as a restriction of a PFA where the symbol and transition probabilities are modeled independently:

\begin{defi}
A hidden Markov model (HMM) is a PFA $\mathcal{A} = \{\Sigma, Q, I, T, F\}$, where in addition to the requirements of a PFA, the transition probability function can be decomposed into two independent symbol $T_{\Sigma} : Q \times \Sigma \to \left[0,1\right]$ and state $T_{Q}: Q \times Q \to \left[0,1\right]$ transition probability functions such that for all $q \in Q, a \in \Sigma, q' \in Q : T(q,a,q') = T_{\Sigma}(q,a)T_{Q}(q,q')$. The probability functions are such that, for all $q \in Q$, $\sum_{a \in \Sigma} T_{\Sigma}(q,a) = 1$ and $\sum_{q' \in Q} T_{Q}(q,q') = 1$.
\end{defi}

Most definitions of HMMs omit the creation of the joint transition probability distribution $T$. They are often also defined without final probability distribution $F$, which is then also removed from the path probability computation. Without $F$, a PFA (or HMM) computes prefix probability functions over $\Sigma^n$, i.e., $\sum_{s \in \Sigma^n} \mathcal{A}(s) = 1$ for all $1 \leq n$. Intuitively, this represents the probability that a trace starts with $s$.
With $F$, it computes a probability distribution over $\Sigma^*$, i.e., $\sum_{s \in \Sigma^*} \mathcal{A}(s) = 1$, representing that a trace starts with $s$ and then ends. Both versions are useful in practice. When modeling software systems, the choice of which model to use typically depends on whether the ending of a trace is determined by the system that generated the data or by the data-gathering process (e.g., when using sliding windows to create traces). In the first case, it is typically better to use final probabilities. In the second case, it is typically better not to use them as they do not provide any information regarding the system.

In deterministic automata, we assume a unique start state $q_0$ exists. Furthermore, given the current system state $q_i$ and the next event $a_{i+1}$, there is a unique next state $q_{i+1}$. Consequently, deterministic automata reach precisely one state at every step of their computation, making the trace probability computation trivial, as there exists only one possible path to compute, resulting in $O(n)$ run-time. 

\begin{defi}
A probabilistic deterministic finite state automaton (PDFA) is a PFA $\mathcal{A} = \{\Sigma, Q, I, T, F\}$, where in addition to the PFA requirements, the initial and transition probability function are deterministic: there exists exactly one $q_0 \in Q$ such that $I(q) > 0$, and for all $q \in Q, a \in \Sigma$ there exists at most one $q' \in Q$ such that $T(q,a,q') > 0$. For convenience, we define the structure of a PDFA using a transition function $\delta : Q \times \Sigma \to Q \cup \{0\}$ such that for all $q \in Q, a \in \Sigma, q' \in Q$, $\delta(q,a) = q'$ if and only if $T(q,a,q') > 0$. When $\delta(q,a) = 0$ for some $q \in Q$ and $a \in \Sigma$, it holds that $T(q,a,q') = 0$ for all $q' \in Q$.
\end{defi}

Since for each state $q$ and symbol $a$ there exists at most one transition to the next state $q'$, PDFAs are typically defined using a symbol probability function $T_{\Sigma} : Q \times \Sigma \to Q$. We use the full transition function $T$ to highlight the restrictions imposed by determinism. Although PDFAs allow for dependence between their symbol and state transition probabilities, they are overall more restrictive than HMMs. But this does not hurt their representational power much: for any $\epsilon$, any probability distribution represented by an HMM $\mathcal{A}$ can be represented by a PDFA $\mathcal{A}'$ such that for any $s$ : $\mathcal{A}'(s) - \epsilon \leq \mathcal{A}(s) \leq \mathcal{A}'(s) + \epsilon$~\cite{dupont2005links}. However, this PDFA can be much larger than the corresponding HMM. 
In practice, PDFAs are more efficient in probability computation and learning. Moreover, determinism can be important when modeling software behavior because, since they can be in exactly one state at each point in time, the resulting models are easier to interpret than their non-deterministic counterparts. For the same reason, non-deterministic models are more robust to noise, as minor sequence deviations affect only a few state paths.

PDFAs and their extensions have frequently been used to model tasks in both software and cyber-physical systems; see, e.g., \cite{nadeem2021alert,klerx2014model,watanabe2021probabilistic,hammerschmidt2016efficient} and \cite{lin2020safety,schmidt2014online,lin2018tabor,maier2014online,pellegrino2017learning}. In addition to their ability to compute probabilities and predict future symbols, their deterministic nature provides insight into a system's inner workings. It can even be used to fully reverse engineer a system \cite{antunes2011reverse}.

\section{State merging in FlexFringe}\label{sec:FlexFringe}
Given a finite data set of example sequences, $D$ called the \emph{input sample}, the goal of PDFA learning (or \emph{identification}) is to find a (non-unique) \emph{small} PDFA $\mathcal{A}$ that is \emph{consistent} with $D$. We call such sequences \emph{positive} or \emph{unlabeled}. In contrast, DFAs are commonly learned from labeled data containing both positive and negative sequences. PDFA size is typically measured by the \emph{number of states} ($|Q|$) or \emph{transitions} ($| \{ (q,a) \in Q \times \Sigma : \delta(q,a) \not= 0 \} |$). Finding a small and consistent (P)DFA is a well-known hard problem and an active research topic in the grammatical inference community; see, e.g.~\cite{higuera2010book}. One of the main methods, originating from the famous RPNI~\cite{oncina1992inferring} (for DFAs) and Alergia~\cite{carrasco1994learning} (for PDFAs) algorithms, is \emph{state merging}. This method starts from a large tree-shaped model that captures the input data exactly. It then iteratively combines (merges) states, resulting in an increasingly smaller model. The method ends when no further merge is possible, i.e., when all possible merges result in an inconsistent model.

When learning DFAs, consistency is typically defined as the model accepting negative sequences or rejecting positive sequences. When learning PDFAs, consistency is tricky to define since only positive data is available. Most methods define consistency by putting restrictions on the merge steps performed by the algorithm. Intuitively, a merge step is consistent when for two states $q$ and $q'$, the future sequences in $D$ occurring after reaching $q$ or $q'$ are similarly distributed. Since different past sequences can reach $q$ and $q'$, this boils down to a type of test for a \emph{Markov property}: the future is independent from the past given the current state, i.e., $\frac{\mathcal{A}(s_qz)}{\mathcal{A}_p(s_q)} \approx \frac{\mathcal{A}(s_{q'}z)}{\mathcal{A}_p(s_{q'})}$ for all suffixes $z \in \Sigma^*$, where $\mathcal{A}_p$ is the prefix probability computed without final probability, and $s_q$ and $s_{q'}$ are prefixes ending in state $q$ and $q'$, respectively. States in a PDFA can thus be thought of as \emph{clusters of prefixes with similarly distributed suffixes}. We call a PDFA $\mathcal{A}$ consistent when this Markov property test succeeds for all prefixes leading to the same state in $\mathcal{A}$, i.e., when $\frac{\mathcal{A}(s_qz)}{\mathcal{A}_p(s_q)} \approx \frac{\mathcal{A}(s_{q}'z)}{\mathcal{A}_p(s_{q}')}$, where $s_q$ and $s_{q}'$ are different prefixes ending in state $q$. This is equivalent to requiring all merges resulting in $\mathcal{A}$ to be consistent.

Intuitively, one could learn a PDFA model by running a clustering algorithm on prefixes with suffix probability differences as distances. When $s_q$ and $s_{q'}$ are clustered together into cluster $c$, however, there can be an $a \in \Sigma$ such that $s_qa$ and $s_{q'}a$ are in different clusters $c_1$ and $c_2$. For the resulting PDFA states, we then have that $T(c,a,c_1) > 0$ and $T(c,a,c_2) > 0$, making the obtained model non-deterministic. We therefore require that for all $z \in \Sigma^*$, if $s_q$ and $s_{q'}$ are clustered, then $s_qz$ and $s_{q'}z$ are clustered as well. This is called the \emph{determinization constraint} and requires performing Markov property tests $\frac{\mathcal{A}(s_qyz)}{\mathcal{A}_p(s_qy)} \approx \frac{\mathcal{A}(s_{q'}yz)}{\mathcal{A}_p(s_{q'}y)}$ on all suffixes $y, z \in \Sigma^*$.
$\mathcal{A}$ and $\mathcal{A}_p$ are unknown during the learning process, but they can be estimated using the partial model constructed by the learning algorithm. We define a merge as consistent when all these Markov property tests are satisfied.

\begin{figure}
    \centering
    \includegraphics[width=\textwidth]{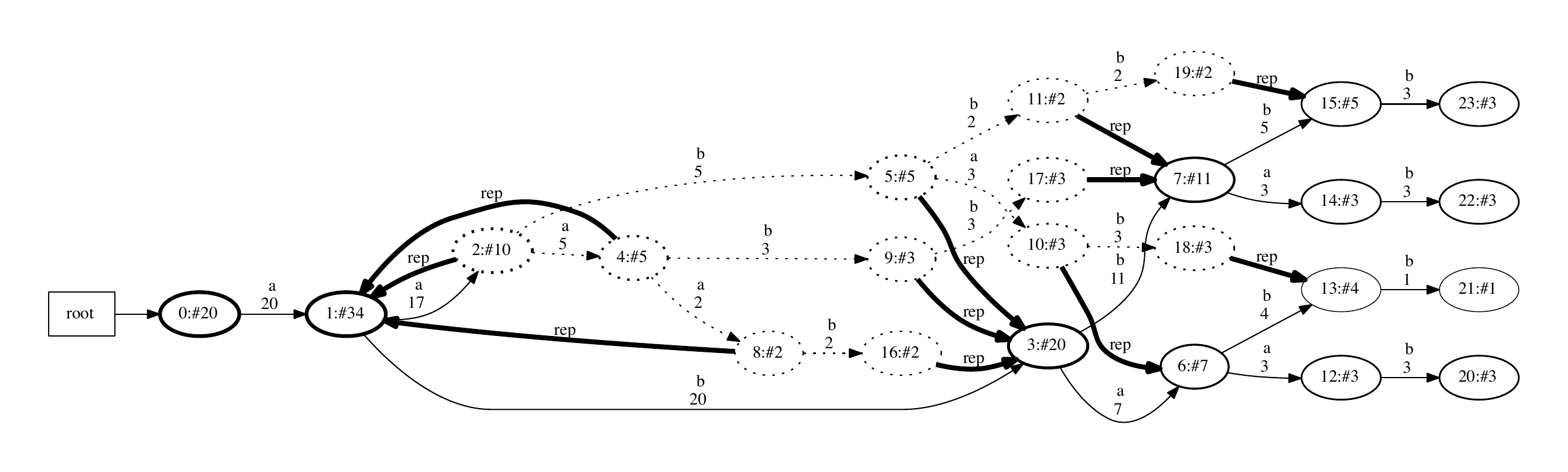}
    \caption{The union/find data structure in FlexFringe after performing the first merge operation (state 2, reached by a-a, is merged with state 1, creating a self-loop). For clarity, we removed the fin and path counts. The unmerged part of the PDFA is solid, and the merged parts are dotted. The arcs labeled ``rep'' are the representative pointers for the union/find structure. Whenever the algorithm queries a state with a representative, the structure follows the ``rep'' pointers until it finds a state without a representative and returns this one instead. Thus, when looking for the target of the transition with label ``a'' from state 1, it will return state 1 (the representative of state 2). States with representatives cannot be merged, but it is possible to merge states that are the representative of others, such as state 3.}
    \label{fig:union/find}
\end{figure}

One of the most successful (P)DFA learning algorithms and an efficient method for performing such tests is evidence-driven state-merging (EDSM) in the red-blue framework~\cite{lang1998results}. FlexFringe implements this framework, using union/find structures to keep track of performed merges and to efficiently undo them, see Figure~\ref{fig:union/find}. Like most state merging methods, FlexFringe first constructs a tree-shaped PDFA $\mathit{A}$ known as a \emph{prefix tree} from the input sample $D$, see Figure~\ref{fig:learning_example}. Afterward, it iteratively merges the states of $\mathit{A}$. Initially, since every prefix leads to a unique state, $\mathit{A}$ is \emph{consistent} with $D$. A \emph{merge} (see Algorithm~\ref{alg:merge}, and Figures~\ref{fig:union/find} and~\ref{fig:red_blue}) of two states $q$ and $q'$ combines the states into one by setting the \emph{representative} variable from the union/find structure of $q'$ to $q$. After this merge, whenever a sequence computation returns $q'$, it returns the representative $q$ instead of $q'$. A merge is only allowed if the states are \emph{consistent}, determined using a statistical test or distance computation based on their future sequences. A powerful feature of FlexFringe is that algorithm users can implement their own test by adding a single file called the \emph{evaluation function} to the source code. For example, using only 100 lines of code, it is possible to add additional attributes such as continuous sensor reading to input symbols and use these in a new statistical test to learn regression automata~\cite{lin2016short}.

When a merge introduces a non-deterministic choice, i.e., both $\delta(q,a)$ and $\delta(q',a)$ are non-zero, the target states of these transitions are merged as well to satisfy the determinization constraint, this is called the \emph{determinization} process (the for loop in Algorithm~\ref{alg:merge}). 
The result of a single merge is a PDFA that is smaller than before (by following the representatives) and still consistent with the input sample $D$ as specified by the evaluation function, since the merge check is performed recursively on all states merged during determinization. FlexFringe repeatedly applies this state merging process until no more consistent merges are possible.

\begin{algorithm}
\caption{Merging two states: {\sf merge} ($\mathcal{A}$, $q$, $q'$) \label{alg:merge}}
\begin{algorithmic}
\REQUIRE PDFA $\mathcal{A}$ and two states $q,q'$ from $\mathcal{A}$
\ENSURE merge $q$ and $q'$ if consistent and return \textsc{true}, return \textsc{false} otherwise
\STATE {\bf if} ${\sf consistent}(q,q')$ is \textsc{false} {\bf return} \textsc{false}
\STATE set ${\sf representative}(q') = q$, update ${\sf score}(q,q')$, add counts of $q'$ to $q$
\FOR{ all $a \in \Sigma$ } 
\IF{ $\delta(q',a) \not= 0$ }
\STATE {\bf if} $\delta(q,a) \not= 0$ call ${\sf merge}(A',\delta(q,a),\delta(q',a))$
\STATE {\bf else} set $\delta(q,a) = \delta(q',a)$
\ENDIF
\ENDFOR
\STATE {\bf if} any merge returned \textsc{false}: \textbf{return} \textsc{false}
\STATE \textbf{else return} \textsc{true}
\end{algorithmic}
\end{algorithm}

\subsection{The red-blue framework}
\begin{figure}[t]
\begin{center}
\vspace{-10pt}
\includegraphics[width=\textwidth]{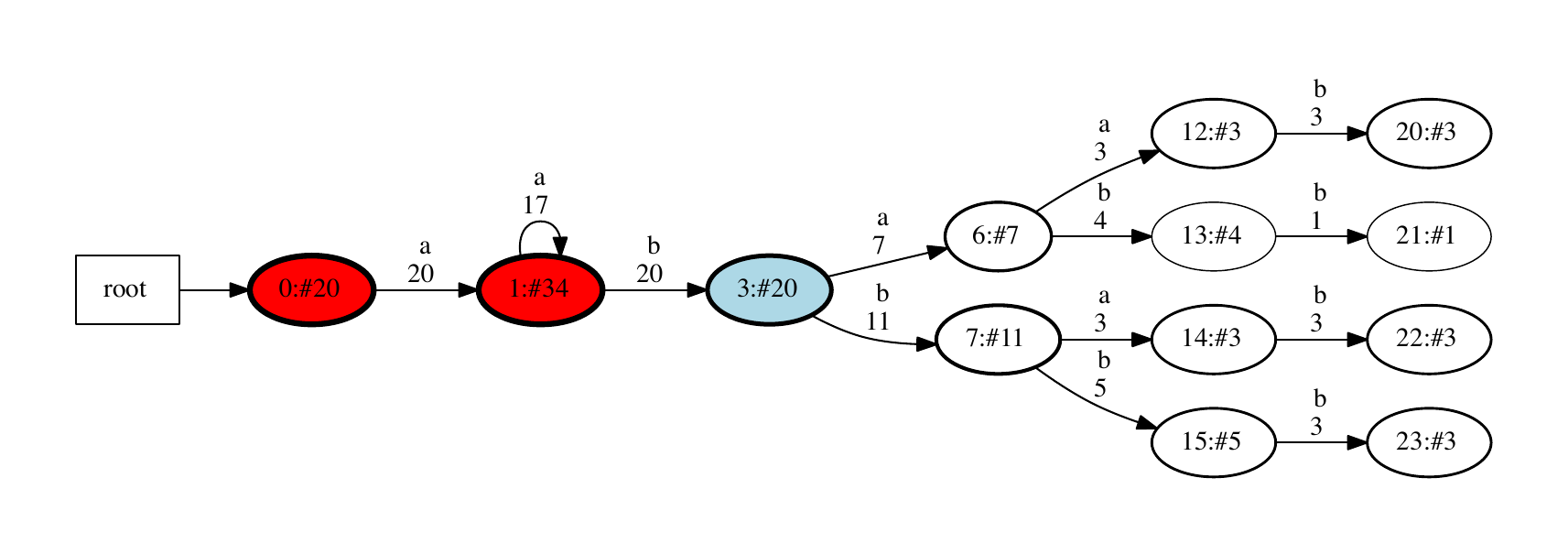}
\caption{The red-blue framework corresponding to the union/find sets from Figure~\ref{fig:union/find}. The red states are the identified parts of the automaton. The blue states are the current candidates for merging. The uncolored states are pieces of the prefix tree that can, at this stage, only be merged during determinization. Currently, only state 3 is a merge candidate. FlexFringe will test the merges of state 3 with 0 and state 3 with 1. If consistent, the highest-scoring merge will be performed. If both are inconsistent, state 3 will be colored red, and states 6 and 7 will be colored blue. }
\label{fig:red_blue}
\end{center}
\end{figure}

The successful \emph{red-blue framework}~\cite{lang1998results} follows the state-merging algorithm just described and adds colors (red and blue) to states to guide the merge process. The framework maintains a core of red states with a fringe of blue states (see Figure~\ref{fig:red_blue} and Algorithm~\ref{alg:red_blue}). A red-blue algorithm performs merges only between blue and red states (although FlexFringe has a parameter for allowing blue-blue merges). When there exists a blue state for which no consistent merge is possible, the algorithm changes the color of this blue state into red, effectively identifying a new state in the PDFA (or DFA) model. The red core of the PDFA can be viewed as a part of the PDFA that is assumed to be correctly identified. Any non-red state $q'$ for which there exists a transition $\delta(q,a) = q'$ from a red state $q$, is colored blue. The blue states are \emph{merge candidates}.

A red-blue state-merging algorithm is complete because it can produce any PDFA that can be obtained by performing merges in the prefix tree. Furthermore, it is more efficient than standard state merging since it considers fewer merges than all pairs of states from the prefix tree. Note that ${\sf undo\_merge}$ (c.f. Algorithm \ref{alg:red_blue}) is highly efficient when using union/find structures because only the representative variables (pointers) need to be reset to their original values and counts updated. Our current implementation does not perform the path compression operations commonly performed in union/find structures. Although path compression is useful to decrease lookup speed, it makes undoing merges much more complex. Moreover, since lookup is not an efficiency bottleneck in FlexFringe, we opted not to use it.

\begin{algorithm}[t]
\caption{State-merging in the red-blue framework\label{alg:red_blue}}
\begin{algorithmic}
\REQUIRE a data set $D$
\ENSURE $\mathcal{A}$ is a small PDFA that is consistent with $D$
\STATE construct the prefix tree $\mathcal{A}$ from $D$
\STATE color the start state of $\mathcal{A}$ red and all of its children blue
\WHILE{ $\mathcal{A}$ contains blue states }
\FOR{ every blue state $q'$ }
\FOR{ every red state $q$ }
\STATE call $\textsc{consistent} = {\sf merge}(\mathcal{A},q,q')$, compute the merge $\textsc{score}$
\STATE call ${\sf undo\_merge}(\mathcal{A},q,q')$
\ENDFOR
\ENDFOR
\IF{ one of the merges is $\textsc{consistent}$ }
\STATE perform the $\textsc{consistent}$ ${\sf
merge}(\mathcal{A},q,q')$ with highest $\textsc{score}$
\ELSE 
 \STATE color the first blue state in a given \textsc{order} red
\ENDIF
 \STATE color all children of red states in $\mathit{A}$ blue
\ENDWHILE
\RETURN $\mathit{A}$
\end{algorithmic}
\end{algorithm}

\subsection{Improvements to red-blue state-merging}

In the grammatical inference community, there has been much research into improving merging in the red-blue framework. Initially, state merging algorithms \cite{oncina1992inferring, carrasco1994learning} used an order that colors shallow states first, i.e., those closer to the start state or root of the prefix tree. In~\cite{lang1999faster}, it was instead suggested to follow an order where the most constrained state is colored first, i.e., the state that has the largest number of inconsistent merges. In FlexFringe, we typically use a largest first order, which colors the most frequent states first. The simple reason for this is that states with large frequency contain more information, and thus, merges are based on more evidence. In our experience, this leads to better automata. One can set these different orders in FlexFringe using the \textsc{shallowfirst} and \textsc{largestblue} input parameters. One can also modify whether FlexFringe should change the color of any blue state without consistent merges to red (parameter \textsc{extend}). These extend operations may violate the merge order, e.g., it colors a non-shallow blue node red when using a shallow-first merge ordering. It is even allowed to consider merges between pairs of blue states (parameter \textsc{blueblue}). How to set these order-influencing parameters is an open problem and highly dependent on the data and use case. In FlexFringe, we provide parameter presets that mimic the original state-merging algorithms from the literature.

In addition to greedily following a predefined order, different search strategies have been studied, such as dependency directed backtracking~\cite{oliveira1998efficient}, using mutually (in)compatible merges~\cite{abela2004mutually}, iterative deepening~\cite{lang1999faster}, beam-search~\cite{bugalho2005inference}, and sat-solving~\cite{heule2010exact}. Most of these have been studied when learning DFA classifiers, i.e., when both positive and negative data are available. 
Most PDFA learning algorithms rely on greedy procedures, some with PAC-bounds that guarantee performance when sufficient data is available~\cite{castro2008towards,clark2004pac,balle2013learning}. There exist some works that minimize Akaike's Information Criterion (AIC) \cite{verwer2010efficient}, or a Minimum Description Length measure \cite{adriaans2007power}. In FlexFringe, a best-first beam-search strategy~\cite{bugalho2005inference} is implemented that minimizes the AIC. We are currently working on implementing more search strategies, including using SAT solvers when learning PDFAs similar to their use for DFAs~\cite{heule2013software}.

\subsection{FlexFringe's parameters}
FlexFringe is built to offer flexibility in the type of merging strategy a user wants to apply. Every dataset is different and can use a different way to perform the basic state-merging algorithm. The core red-blue state-merging functionality (Algorithm~\ref{alg:red_blue}) cannot be changed. However, many parameters can be used to change how this framework is applied. We list the most important ones in Table~\ref{tab:parameters}. These are discussed in more detail in different sections of this paper.

\begin{center}
\begin{table}[h]
    \centering
    {\small
    \begin{tabular}{c|c|l} 
       \textsc{largestblue} & Bool & when set to true, only the most frequent blue state is considered\\
       \textsc{shallowfirst} & Bool & consider coloring blue states closest to the root first\\   
       \textsc{extend} & Bool & extend (color red) blue states that cannot be merged with any red\\
       \textsc{blueblue} & Bool & whether to allow merges between pairs of blue states\\
       \textsc{redfixed} & Bool & merges that add new transitions to red states are inconsistent\\
       \textsc{markovian} & Bool & states with different incoming transition symbols are inconsistent\\
       \textsc{ktail} & Int & only perform merge consistency checks up to \textsc{ktail} depth \\
       \hline
       \textsc{sinkson}  & Bool & whether to use sink states (unmergeable blue states)\\
       \textsc{sink\_count}  & Int & states with smaller frequency counts are sinks \\
       \textsc{state\_count} & Int & the minimum count for a state to be used in consistency tests\\
       \textsc{symbol\_count} & Int & the minimum count for a symbol to be used in consistency tests\\
       \textsc{correction} & Float & Laplace smoothing addition to every symbol count after pooling\\
       \textsc{finalprob} & Bool & whether to use final probabilities in the model and merge tests\\
       \textsc{confidence\_bound} & Float & the input parameter used by (statistical) consistency checks\\
       \hline
       \textsc{mode} & Text & whether to use greedy, search, or predict mode (see Section~\ref{sec:run})\\
       \textsc{apta\_file} & Text & a path to a learned model file for use in predict mode
    \end{tabular}
    }
    \caption{A list of parameters that influence the basic operation of FlexFringe and their meaning. These are discussed in more detail in Sections 3.2, 5.1-5.4., and 6}
    \label{tab:parameters}
\end{table}
\end{center}

\section{Implemented evaluation functions}
\label{sec:functions}
Given a potential merge state pair $(q, q')$, an evaluation function has to implement a \emph{consistency check} and a \emph{score}. As mentioned above, the consistency check determines whether a merge is possible. The score evaluation determines which merge to perform from a set of possible ones. For example, when learning a PDFA, merge consistency can be determined by comparing the distance of suffix distributions of two states to a threshold. When several pairs of states have a distance below this threshold, it makes sense to first merge the states with the smallest distance. In FlexFringe, both of these functions are defined by the user. The score and consistency checks are often computed using the same quantity, such as a distance, but this is not required. For learning PDFAs, FlexFringe provides several well-known consistency checks, which we describe below, along with their score computation. Note that these evaluation functions are applied to all pairs of states merged during determinization (the recursion in  Algorithm~\ref{alg:merge}). In other words, each merge calls these functions multiple (sometimes hundreds of) times, depending on the size of the prefix tree. It is, therefore, important that they can be computed quickly. For instance, FlexFringe computes counts incrementally and only updates them for the states that get merged during the merge and determinization procedure.

\subsection{Counts}
The currently implemented consistency checks for learning PDFAs use state and symbol counts to determine consistency. These are maintained by FlexFringe in each state using standard maps. Every time a merge is done or undone, the counts in these maps are updated. Since this has to be done for each pair of states merged during determinization, this is typically the most run-time intensive part of the merge procedure. FlexFringe, therefore, only performs such updates when the counts are used. For instance, they are not maintained when learning deterministic automata using the evidence-driven state-merging algorithm. To implement the checks for learning PDFAs, FlexFringe maintains counts that are used to compute the following:

\begin{itemize}
\item $C(q)$: the frequency count of state $q$.
\item $C(q,a)$: the frequency count of symbol $a$ in state $q$.
\item $T_{\Sigma}(q,a) = \frac{C(q,a)}{C(q)}$: the probability of symbol $a$ in state $q$.
\item $F(q) = 1.0 - \sum_{a \in \Sigma} T_{\Sigma}(q,a)$: the final probability of state $q$.
\item $|\mathcal{A}|$: the number of transitions in PDFA $\mathcal{A}$.
\end{itemize}

\subsection{Alergia}

Alergia \cite{carrasco1994learning} is one of the first and still a very successful algorithm for learning PDFAs. It relies on a test derived from the Hoeffding bound to determine merge consistency checks. For each potential merge pair $(q,q')$, it tests for all $a \in \Sigma$ whether

\[
\left| T_{\Sigma}(q,a) - T_{\Sigma}(q',a) \right| < \sqrt{\frac{1}{2} \ln \frac{2}{\alpha}} \left( \frac{1}{\sqrt{C(q)}} + \frac{1}{\sqrt{C(q')}} \right) 
\]

where $\alpha$ is a user-defined parameter set in FlexFringe using the \textsc{confidence\_bound} parameter. When using final probabilities, the final probability function $F$ is used in the same way as the symbol probability function $S$ (this holds for all evaluation functions), i.e., it then also tests whether:

\[
\left| F(q) - F(q') \right| < \sqrt{\frac{1}{2} \ln \frac{2}{\alpha}} \left( \frac{1}{\sqrt{C(q)}} + \frac{1}{\sqrt{C(q')}} \right) 
\]

The Alergia check guarantees that the outgoing symbol distributions are not significantly different for every pair of merged states. In the original Alergia paper, the state merging algorithm does not implement a score. Instead, it defines a shallow first search order and iteratively performs the first consistent merge in this order. In FlexFringe, Alergia is implemented with the default score of summing up the differences between the left-hand and right-hand sides of the above equation over all pairs of tested states. Let $M \subseteq Q \times Q$ be the set of pairs of states that are checked during merge determinization, i.e., all the $q$ and $q'$ pairs provided as input to Algorithm~\ref{alg:merge} during the recursive $\textsf{merge}$ calls. The Alergia score is:

\[
\sum_{(q,q') \in M}\left( \sqrt{\frac{1}{2} \ln \frac{2}{\alpha}} \left( \frac{1}{\sqrt{C(q)}} + \frac{1}{\sqrt{C(q')}} \right) - \left| T_{\Sigma}(q,a) - T_{\Sigma}(q',a) \right| \right)
\]

When using final probabilities, this score is also computed for final probabilities. The Alergia score becomes larger when the differences between the distributions become smaller. In addition, this score prefers merges that perform many merges during determinization. The underlying intuition is that a consistency check is performed for every performed merge. Hence, we are more confident of the consistency of merges that merge more states.

\subsection{Likelihood-ratio}

A likelihood ratio test is introduced in ~\cite{verwer2010likelihood} to overcome a possible weakness of Alergia. In Alergia, each pair of merged states is tested independently. When determinization merges hundreds of states, we should not be surprised that several of these independent tests fail. This prevents states from merging, resulting in a larger PDFA. The likelihood ratio test aims to overcome this by computing a single test for the entire merge procedure, including determinization. It compares the PDFA before the merge $\mathcal{A}$ to the PDFA after the merge $\mathcal{A}'$, computing their log-likelihood and the number of parameters of the PDFA. The log-likelihood of a PDFA is simply the log of all probabilities it assigns to the training data $\sum_{s \in D} \log (\mathcal{A}(s))$. The number of parameters (or degrees of freedom) of a PDFA is the number of transitions since each requires a probability to be estimated. We use $|\mathcal{A}|$ to denote the number of transitions of PDFA $\mathcal{A}$. 
Because the two models are nested ($\mathcal{A}'$ is a restriction/grouping of $\mathcal{A}$), we can compute a likelihood ratio test to determine whether parameter reduction (fewer transitions) outweighs the decrease in likelihood. When it does, a merge is considered consistent. The function it computes is:

\[
1.0 - \chi^2 \left( 2 \sum_{q \in Q, a \in \Sigma} C(q,a) \log(T_{\Sigma}(q,a)) - 2 \sum_{q' \in Q', a \in \Sigma} C(q',a) \log(S'(q',a)), |\mathcal{A}| - |\mathcal{A}'| \right) > \alpha
\]

where $\alpha$ is a user-defined parameter (\textsc{confidence\_bound}), $\chi^2 (v,n)$ is the value of $v$ in the chi-squared distribution with $n$ degrees of freedom. In the case that $C(q,a)$ equals $0$, $C(q,a) \log(T_{\Sigma}(q,a))$ is set to $0$. In FlexFringe, this function is computed incrementally by tracing which parameters get removed during a merge and their effect on the log likelihood. As a score, likelihood ratio uses the p-value obtained from the $\chi^2$ function, i.e.:

\[
1.0 - \chi^2 \left( 2 \sum_{q \in Q, a \in \Sigma} C(q,a) \log(T_{\Sigma}(q,a)) - 2 \sum_{q' \in Q', a \in \Sigma} C(q',a) \log(S'(q',a)), |\mathcal{A}| - |\mathcal{A}'| \right)
\]

A larger score indicates that the decrease in likelihood is less significant, i.e., that the distributions modeled by $\mathcal{A}$ and $\mathcal{A}'$ are more similar.

\subsection{MDI}

The MDI algorithm~\cite{thollard2000probabilistic} is an earlier approach to overcome possible weaknesses of Alergia, mainly that there is no way to bound the distance of the learned PDFA from the data sample. Like the likelihood ratio, MDI computes the likelihood and the number of parameters. Instead of comparing these directly using a test, MDI uses them to compute the Kullback-Leibler divergence from the models before merging $\mathcal{A}$ and after merging $\mathcal{A}'$ to the distribution in the original data sample $D$. The distribution of $D$ is determined using the prefix tree $\mathcal{A}_p$. When a merge makes this distance too large, it is considered inconsistent:

\[
\frac{ \left( \sum_{q \in Q_p, a \in \Sigma} C_p(q,a) S_p(q,a) \cdot \left( \log(T_{\Sigma}(\hat{q},a)) - \log(S'(\hat{q}',a)) \right) \right)}{|\mathcal{A}| - |\mathcal{A}'|} < \alpha
\]

where $C_p(q,a)$ and $S_p(q,a)$ is the count information from the prefix tree $\mathcal{A}_p$, $\hat{q}$ and $\hat{q}'$ are the states that are merged with $q$ in $\mathcal{A}$ and $\mathcal{A}'$ respectively. As before, and $\alpha$ is a user-defined parameter (\textsc{confidence\_bound}). As a score, MDI in FlexFringe uses the negative increase in Kullback-Leibler divergence:

\[
\frac{ -\left( \sum_{q \in Q_p, a \in \Sigma} C_p(q,a) S_p(q,a) \cdot \left( \log(T_{\Sigma}(\hat{q},a)) - \log(S'(\hat{q}',a)) \right) \right)}{|\mathcal{A}| - |\mathcal{A}'|}
\]

A larger score corresponds to the model's distribution staying closer to the distribution of the prefix tree. For efficiency reasons, our implementation differs slightly from the original formulation in~\cite{thollard2000probabilistic}. We use the counts from $D$ to compute the Kullback-Leibler divergence (similar to the likelihood ratio) instead of computing it directly between the different models. Unfortunately, it is currently not possible to perform a run of the original MDI algorithm, which used an exact algorithm to compute the divergence, as this would require too much run-time to be used inside a merge check. This is, however, the only difference between MDI and its implementation in FlexFringe.

\subsection{AIC}

The AIC or Akaike's Information Criterion is a commonly used measure for evaluating probabilistic models ~\cite{akaike1974new}. It is a simple yet effective model selection method for making the trade-off between the number of parameters and likelihood. It is similar to the likelihood ratio function but does not rely on the $\chi^2$ function. It simply aims to minimize the number of parameters minus the log likelihood. In~\cite{verwer2010efficient}, this was used to learn probabilistic real-time automata, similar to using the minimum description length principle for learning DFAs~\cite{adriaans2006using}. In FlexFringe, we simply consider all merges that decrease the AIC as consistent:

\[
2 \left( |\mathcal{A}| - |\mathcal{A}'| \right) - 2 \left(\sum_{q \in Q, a \in \Sigma} C(q,a) \log(T_{\Sigma}(q,a)) - \sum_{q' \in Q', a \in \Sigma} C(q',a) \log(S'(q',a)) \right) > 0
\]

Intuitively, this measures whether the parameter reduction from $\mathcal{A}$ to $\mathcal{A}'$ is greater than the decrease in log likelihood. FlexFringe computes this incrementally. As a score, FlexFringe directly uses the AIC value: 

\[
2 \left( |\mathcal{A}| - |\mathcal{A}'| \right) - 2 \left(\sum_{q \in Q, a \in \Sigma} C(q,a) \log(T_{\Sigma}(q,a)) - \sum_{q' \in Q', a \in \Sigma} C(q',a) \log(S'(q',a)) \right)
\]

\section{Improvements in Speed and Performance}
\label{sec:improvements}
In addition to its efficient implementation and flexibility, FlexFringe introduces several techniques that improve run-time and state-merging performance. 
The above evaluation functions work for states that are ``sufficiently frequent''. When merging infrequent states, however, they can give bad performance. For instance, Alergia will nearly always merge infrequent states since they will never provide sufficient evidence to determine an inconsistency. As a result, these merges are somewhat arbitrary and can hurt both performance and the insight one can get from the learned models. FlexFringe, therefore, implements several techniques that deal with low-frequency states and transitions.

\subsection{Sinks}
Sinks are states satisfying user-defined conditions that are ignored by the merging algorithm. The idea of using sinks originated from DFASAT in the Stamina challenge \cite{walkinshaw2013stamina,heule2013software}. Data was labeled in the competition, and a garbage state was needed to model states only reached by negative sequences. A garbage state is a state that rejects all traces that reach it, independent of their future execution. Keeping such states as merge candidates would allow these to be merged with the rest of the automaton, thereby combining negative and positive sequences, which can only lower performance. Sinks were introduced to explicitly model such garbage states.

In PDFAs, the default condition defines sinks as states that are reached less than \textsc{sink\_count} (a user-defined parameter) times by sequences from the input data $D$. It is possible to extend this with other conditions for different applications, such as a connection closing when learning a communication protocol~\cite{de2015protocol}. In FlexFringe's merging routines, sinks are never considered as merge candidates, i.e., blue states that are sinks are ignored. They are, however, merged normally during determinization. Since the counts from merged states are combined, sinks can become more frequent and thus become a merge candidate in a subsequent iteration. The merging routines continue until all remaining merge candidates are sinks. By default, these sinks and their future states are not printed as output to the automaton model. The transitions to sink states, the sink states themselves, and all subsequent states are not printed. This can be modified using several input parameters, including printing them to the output file or merging them with red states or only with other sinks after the greedy merge process (Algorithm~\ref{alg:red_blue}) ends.

\subsection{Pooling}
Frequency pooling is a common technique to improve the reliability of statistical tests when faced with infrequent symbols/events/bins. The idea is to combine the frequency of infrequent symbols and thus gain confidence in the outcome of statistical tests. When learning PDFAs, frequency pooling is essential as most states merged during determinization are infrequent. Every blue state considered for merging is the root of a prefix tree with frequent states near the root and infrequent states in all of its branches. Pooling can be pretty straightforward. For example, one can combine the counts of symbols that occur under a user-defined threshold in either or both states of the merge-pair.

However, we noticed that straightforward pooling strategies, such as combining columns containing cells with a count of less than 5, can miss obvious differences. Consider, for instance, the following contingency table of symbol counts for two states $q$ and $q'$:

\begin{center}
\begin{tabular}{l|ccccc||c}
      ~   & a & b & c & d & e & pool \\ \hline
    $q$  & 8 & 8 & 4 & 0 & 0 & 20 \\
    $q'$ & 0 & 0 & 4 & 8 & 8 & 20 \\
\end{tabular} 
\end{center}

Although states $q$ and $q'$ show different distributions (only 4 occurrences overlap in both states), the pooled counts are identical. This creates problems for state merging as the algorithm will consider such merges consistent while they are not. Merging these small counts can greatly affect the final model since they are added to the representative state $q$ (see Algorithm~\ref{alg:merge}). This thus influences all subsequent statistical tests performed on $q$. This effect can be large since many states have small counts in the prefix tree. In FlexFringe, we opted for a pooling strategy that aims not to hide such differences. We build two pools:

\begin{center}
\begin{tabular}{l|ccccc||cc}
      ~   & a & b & c & d & e & pool1 & pool2 \\ \hline
    $q$  & 8 & 8 & 4 & 0 & 0 & 20 & 4 \\
    $q'$ & 0 & 0 & 4 & 8 & 8 & 4 & 20 \\
\end{tabular} 
\end{center}

The first pool sums the frequency counts for all symbols that occur less than threshold 5 in $q$ ($d, \ldots, h$), and the second those in $q'$ ($a, \ldots, e$). The counts that occur infrequently in both states are added to both pools. The parameter in FlexFringe used for this threshold setting is \textsc{symbol\_count}. Note that pooling occurs only when computing a consistency check or score. It is independent of the consideration of low-count states as sink states. When learning a PDFA model for a software system, it can make sense to set \textsc{symbol\_count} to 1. This causes the merging process to be more influenced by the absence/presence of symbols, which, in our experience, is useful when modeling software processes. This way, it is possible to detect differences between states with many different occurring symbols, but each only occurs a few times, for example:

\begin{center}
\begin{tabular}{l|cccccc||cc}
      ~   & $a_1$ & $\ldots$ & $a_m$ & $a_{m+1}$ & $\ldots$ & $a_n$ & pool1 & pool2 \\ \hline
    $q$  & 1 & $\ldots$ & 1 & 0 & $\ldots$ & 0 & $m$ & 0 \\
    $q'$ & 0 & $\ldots$ & 0 & 1 & $\ldots$ & 1 & 0 & $n-m$ \\
\end{tabular} 
\end{center}

The statistical tests in FlexFringe ignore states with an occurrence frequency lower than \textsc{state\_count}. This means that whenever one of the two states being merged is infrequent during determinization, it still performs the merge but does not compute a statistical test. It is therefore important to set \textsc{sink\_count} greater than \textsc{state\_count}, as otherwise states will be considered as merge candidates that are too infrequent to compute any statistical test. Removing infrequent states from the statistical test computation is especially important when using the likelihood ratio test, as it can behave incorrectly when merging many infrequent states. Using the \textsc{state\_count} parameter, these counts are not added to the likelihood value, and the size reduction (number of parameters) is also not considered. FlexFringe also uses a Laplace smoothing by adding \textsc{correction} counts (default 1) to every frequency count after pooling.

\subsection{Counting parameters}
Many evaluation functions require counting statistical parameters before and after a merge. A statistical parameter is a stochastic variable that is estimated from data, such as the sample mean or variance. For a PDFA $\mathcal{A} = \{\Sigma, Q, I, T, F\}$, the data are used to estimate the probability values of $T$ and $F$. The number of parameters is thus equal to the number of such values, i.e., the sizes of $T$ and $F$. A problem is computing the reduction in the number of parameters when merging states. Because a merge can combine many states during determinization, counting one parameter more or less for each state greatly influences the resulting automaton model.

Each state contains statistical parameters for estimating $T$ and $F$. It contains one parameter for every possible symbol, plus one extra for the final probability. It makes little sense to include parameters for symbols that do not occur, i.e., symbols $a$ for which $\delta(q,a) = 0$. Counting parameters for these would give a huge preference to merging, as every pair of merged states reduces this amount by the size of the alphabet $|\Sigma|$. 
Instead, we count an additional parameter only for symbols with non-zero counts in both states before merging. In other words, we count every transition as a parameter. We thus measure the size of a PDFA by counting the number of transitions instead of the number of states.

\subsection{Merge constraints}

In addition to ways to deal with low-frequency counts and symbol sparsity, FlexFringe contains several parameters that influence which merges are considered. For PDFAs, one of the most important parameters is \textsc{largestblue}. When set to true, FlexFringe only considers merges with the most frequently occurring blue state. This dramatically reduces run-time because instead of trying all possible red-blue merge pairs (quadratic), it only considers merges between all red states with a single blue state (linear). It typically also improves performance, as merging the most frequent states first makes sense when testing for consistency using statistical tests. The \textsc{finalprob} parameter is also important. When set to true, it ensures FlexFringe models final probabilities $F$, thus learning distributions over $\Sigma^*$ instead of over $\Sigma^n$. When set to false, it sets $F(q) = 0$ for every $q \in Q$. Final probabilities should only be used when the ending of sequences contains information, e.g., not when learning from sliding windows that start and end arbitrarily.

Several other parameters can be useful when learning PDFAs. Firstly, \textsc{redfixed} makes sure that merges cannot add new transitions to red states; when they do, they are considered inconsistent. The key idea is that the red states are already learned/identified, so we should not modify their structure. We still allow modifications to their symbol and final probabilities due to increasing occurrence counts.

The \textsc{blueblue} parameter allows merges between pairs of blue states in addition to red-blue merges. Although state merging in the red-blue framework is complete in the sense that it can return any possible automaton, sometimes it can force a merge with a low evidence score. Allowing blue-blue merge pairs can avoid such merges. 

The \textsc{markovian} parameter creates a Markov-like constraint. It disallows merges between states with different incoming transition labels when set to $1$. When set to $2$ (or $3$, ...), it also requires their parents, i.e., past states, (and their parents, ...) to have the same incoming label. When running likelihood ratio with a very low statistical test threshold (or negative), and \textsc{markovian} set to 1, it creates a Markov chain. With a larger setting, it creates an n-Gram model. Combined with one of the statistical consistency checks described above, it creates a deterministic version of a labeled Markov chain~\cite{alur1991model}.

FlexFringe also implements the well-known kTails algorithm, which is often used in software engineering~\cite{biermann1972synthesis}, i.e., only taking futures sequences up to length $k$ into account, which can be accessed using the \textsc{ktail} parameter.

\subsection{Searching}
Much of the efficiency in FlexFringe is achieved using a union/find data structure, which allows to perform and undo merges quickly. Most of the time is typically spent reading, writing, and updating the data structures used to compute a merge's consistency and score. Doing and undoing merges typically is not a performance bottleneck. Search routines can, therefore, quickly switch to different merge paths (undoing and redoing merges) by only performing union/find updates. In this way, FlexFringe can try to optimize a global objective. For PDFAs, it minimizes the AIC of the resulting model. For this purpose, we have implemented a simple best-first search strategy similar to ed-beam~\cite{bugalho2005inference}. You can run this by setting the \textsc{mode} parameter to the \textsc{search} setting. When set to search, FlexFringe prints out an overview of the search space size. Whenever it finds a new model that improves the AIC, it prints the new model to a file.

\section{An example run}\label{sec:run}
We give an example run of FlexFringe that demonstrates its ease of use and how to use the output provided by FlexFringe to decide on parameter values. We run FlexFringe from the command line using Alergia on the data from Figure~\ref{fig:learning_example}.

\begin{verbatim}
./flexfringe --ini ini/alergia.ini test_paper.dat
\end{verbatim}

The first argument gives the initialization file (alergia.ini), which contains parameters for running the Alergia algorithm:

\begin{verbatim}[default]
heuristic-name = alergia
data-name = alergia_data
confidence_bound = 0.95
largestblue = 1
finalprob = 1
\end{verbatim}

This specifies which class files to use for evaluation and data processing (alergia). Specifying this at run-time makes switching to a different underlying algorithm easy. Moreover, people interested in developing their own evaluation function heuristic can do so by adding only a single file to the code base. We set Alergia to use \textsc{FINALPROB} and \textsc{LARGESTBLUE}. The \textsc{confidence\_bound} ($\alpha$ parameter used in the consistency check) is set to 0.95, much higher than the default of 0.01. This makes it possible to learn models when presented with only 20 traces. With its default setting, FlexFringe cannot distinguish states from each other and learns a single-state automaton when given such few traces.

The second argument to FlexFringe provides the training data as input (the data from Figure~\ref{fig:learning_example}). Executing the above call to FlexFringe provides the following output:

\begin{verbatim}
Using heuristic alergia
Creating apta using evaluation class alergia
batch mode selected
starting greedy merging
 x20  m5.0184  x20  m2.45255  m2.08965 no more possible merges
deleted merger
\end{verbatim}

Flexfringe first prints checks for the selected evaluation function and search strategy. After that, it outputs information on every performed merge and core extension (coloring a node red). The first output x20 means it extends the core with a state with frequency 20. The second output m5.0184 means it performed a merge with score 5.0184 (summed differences between left-hand side and right-hand side of the performed Alergia tests, see Section~\ref{sec:functions}). This output can be made more verbose, it then also prints the state numbers. It ends when no more merges can be performed and prints whether it successfully freed all allocated memory.

When using FlexFringe, the output can be useful in guiding parameter settings. For instance, merges can be performed with very low scores (printing m0). Such merges can be caused by merging low-frequency states or an incorrect merge order. One can try to avoid them by changing parameters, such as the use of sinks. 

As output, FlexFringe provides two files:

\begin{verbatim}
test_paper.dat.ff.final.dot
test_paper.dat.ff.final.json
\end{verbatim}

The dot formatted file is for visualization purposes only. Graphviz dot\footnote{https://graphviz.org} produces Figure~\ref{fig:learning_example_result} when given this as input. The JSON file can be used for further processing by FlexFringe to make predictions. This is done by calling FlexFringe in predict mode:

\begin{verbatim}
./flexfringe --ini ini/alergia.ini test_paper_test.dat --mode=predict
--aptafile=test_paper.dat.ff.final.json --correction=0
\end{verbatim}

This runs the predict function with the PDFA specified by the JSON input on the data argument. We provide an additional argument that turns off Laplace smoothing, overriding the ini file and simplifying the probability calculations. We run it on a small test containing only the trace ``a b a b a''. This produces a CSV-formatted file as output containing:

\begin{verbatim}
state sequence; score sequence
[1,2,1,2,1,1]; [0,-0.510826,-1.60944,-0.510826,-1.60944,-inf]
\end{verbatim}

As shown in Figure~\ref{fig:learning_example}, the state sequence corresponds to the sequence of states visited by the trace "a b a b a". At the end, it contains state number 1 twice to denote the state the trace ends in. The scores are the log probabilities with base $e$ for each state-symbol combination. The trace always starts with an "a" (log-probability 0). Afterward, it gives the log-probability of producing a "b" symbol in state 1:

\[
\ln(\frac{C(1,b)}{C(q)}) = \ln(\frac{30}{50}) \approx -0.51
\]

The score sequence ends with "-inf" since it tries to compute the log probability of ending in state 1. Without smoothing (\textsc{CORRECTION}), this probability is 0, giving an infinite negative score. Thus, the trace "a b a b a" can be labeled as an anomaly.

FlexFringe comes with a Python wrapper for making the above calls for convenience. This makes it easier to integrate FlexFringe into existing Python data processing pipelines. Moreover, in addition to being open-source, FlexFringe contains pre-compiled binaries for Windows, Mac, and Linux. Tutorials on setting up and using FlexFringe are also available online in the FlexFringe repository.\footnote{https://github.com/tudelft-cda-lab/FlexFringe}

\section{Results on PAutomaC}
\label{sec:results}
To demonstrate the value of the improvements made to general state merging algorithms in FlexFringe, we run each evaluation function on the PAutomaC problem set and compare FlexFringe's performance to the competition winners. PAutomaC was a competition on learning probability distributions over sequences held in 2012~\cite{verwer2014pautomac}. In the competition data, there are 48 data sets with varying properties, such as the type of automaton/model used to generate the sequences (PFA, HMM, or PDFA), the size of the alphabet, and the sparsity/density of transitions. Since deterministic automata can approximate non-deterministic automata~\cite{dupont2005links}, algorithms for deterministic models (such as FlexFringe) can be used for non-deterministic target models. However, one key finding from the PAutomaC competition was that algorithms for learning non-deterministic models perform better when the target is non-deterministic. For evaluation, a test set of unique traces was provided. The task was to assign probabilities to these traces. The assigned probabilities were compared to the ground truth (probabilities assigned by the model that generated the data) using a perplexity metric:

$$Score(C, T, D)=2^{- \sum_{x \in D} P_T(x) * log (P_C(x))  }$$

where $C$ is a submitted candidate model, $T$ is the target model, $D$ is the data set of testing sequences, $P_T(x)$ is the normalized probability of $x$ in the target and $P_C(x)$ is the normalized candidate probability for $x$ submitted by the participant. The perplexity score measures how well the assigned probabilities matched the target probabilities assigned by the ground truth model.

To avoid 0 probabilities in $P_T(X)$, we use Laplace smoothing with a correction of 1. We compare the performance of FlexFringe using different heuristics and parameters to the PAutomaC winner (a Gibbs sampler by Shibata-Yoshinaka that learns a PNFA) and the best-performing method on data generated using PDFAs (a state merging method by team Llorens that learns a PDFA). We obtained the scores for Shibata-Yoshinaka and team Llorens from the PAutomaC competition paper~\cite{verwer2014pautomac}. We first demonstrate the effectiveness of sinks, low-frequency counts, and other improvements using Alergia.

\begin{table}
\centering
\begin{footnotesize}
\begin{tabular}{|l|c|c|ccccccc|}\hline
Nr & Model & Solution & Shibata  & Llorens & Alergia94 & Alergia+ & Likelihood & MDI & AIC \\ \hline
6 & PDFA & 66.99 & 67.01 &  {\bf 67.00} &
        74.05 & 67.01 &
        {\bf 67.00} & 67.54 & {\bf 67.00} \\ \hline
7 & PDFA & 51.22 &  51.25 & 51.26 &
        82.92 & {\bf 51.24} &
        {\bf 51.24} & 51.46 & {\bf 51.24} \\ \hline
9 & PDFA & 20.84 & 20.86 & {\bf 20.85} &
        22.22 & {\bf 20.85} &
        {\bf 20.85} & 20.99 & {\bf 20.85} \\ \hline
11 & PDFA & 31.81 & 31.85 &  32.55 &
        76.53 & {\bf 31.84} &
        31.85 &  {33.56} & {\bf 31.84} \\ \hline
13 & PDFA & 62.81 & {\bf 62.82} & {\bf 62.82} &
        65.01 &  {64.76} &
         {64.86} & 62.87 & {\bf 62.82} \\ \hline
16 & PDFA & 30.71 & {\bf 30.72} &  {\bf 30.72} &
        33.49 & {\bf 30.72} &
        {\bf 30.72} & 30.78 & {\bf 30.72} \\ \hline
18 & PDFA & 57.33 & {\bf 57.33} &  {\bf 57.33} &
        67.04 & {\bf 57.33} &
        {\bf 57.33} & 57.39 & {\bf 57.33} \\ \hline
24 & PDFA & 38.73 & {\bf 38.73} & {\bf 38.73} &
        39.63 & {\bf 38.73} &
        {\bf 38.73} & 38.91 & {\bf 38.73} \\ \hline
26 & PDFA & 80.74 & {\bf 80.83} & 80.84 &
        112.01 & 80.89 &
        80.91 &  {83.52} & 80.98 \\ \hline
27 & PDFA & 42.43 & {\bf 42.46} & {\bf 42.46} &
        80.52 & {\bf 42.46} &
        {\bf 42.46} & 43.49 & 42.47 \\ \hline
32 & PDFA & 32.61 & {\bf 32.62} & {\bf 32.62} &
        33.28 & {\bf 32.62} &
        {\bf 32.62} & 32.65 & {\bf 32.62} \\ \hline
35 & PDFA & 33.78 & {\bf 33.80} & 34.30 &
        72.29 & {\bf 33.80} &
        {\bf 33.80} & 36.81 & 33.81 \\ \hline
40 & PDFA & 8.20 & {\bf 8.21} & {\bf 8.21} &
        9.66 & 8.26 &
        8.67 & 8.52 & 8.23 \\ \hline
42 & PDFA & 16.00 & {\bf 16.01} & {\bf 16.01} &
        16.14 & {\bf 16.01} &
        {\bf 16.01} & 16.05 & {\bf 16.01} \\ \hline
47 & PDFA & 4.119 & {\bf 4.12} & {\bf 4.12} &
        4.65 & {\bf 4.12} &
        {\bf 4.12} & 4.13 & {\bf 4.12} \\ \hline
48 & PDFA & 8.04 & {\bf 8.04} & 8.19 &
        11.73 & {\bf 8.04} &
        {\bf 8.04} & 8.24 & {\bf 8.04} \\ \hline \hline
1 & HMM & 29.90 & {\bf 29.99}  & 30.40 &        
        34.01 &  {31.98} &
         {31.58} &  {31.20} &  {31.19} \\ \hline
2 & HMM & 168.33 & 168.43 & {\bf 168.42} &
        171.21 & 168.43 &
        168.43 & 168.96 & 168.43 \\ \hline
5 & HMM & 33.24 & {\bf 33.24} & {\bf 33.24} &
        34.65 & {\bf 33.24} &
        {\bf 33.24} & 33.31 & {\bf 33.24} \\ \hline
14 & HMM & 116.79 & {\bf 116.84} & {\bf 116.84} &
        117.88 & {\bf 116.84} &
        116.85 & 117.13 & 116.85 \\ \hline
19 & HMM & 17.88 & {\bf 17.88} & 17.92 &
        18.60 & 17.97 &
        17.98 & 17.92 & 17.92 \\ \hline
20 & HMM & 90.97 & {\bf 91.00} &  93.50 &
        149.44 &  {92.36} &
        91.86 &  {98.61} & 91.68 \\ \hline
21 & HMM & 30.52 & {\bf 30.57} & 32.22 &
        83.40 &  {35.25} &
         {35.47} &  {37.31} &  {33.52}\\ \hline   
23 & HMM & 18.41 & {\bf 18.41} & 18.45 &
        18.84 & 18.49 &
        18.44 & 18.47 & 18.45 \\ \hline
25 & HMM & 65.74 & {\bf 65.78} &  67.27 &
        101.97 &  {67.26} &
         {68.24} & 66.83 & 66.96 \\ \hline
28 & HMM & 52.74 & {\bf 52.84} & 53.20 &
        60.83 & 53.77 &
        53.05 & 53.55 & 53.02 \\ \hline
33 & HMM & 31.87 & {\bf 31.87} &  32.03 &
        32.21 & 31.96 &
        31.95 & 32.64 & 31.97 \\ \hline
36 & HMM & 37.99 & {\bf 38.02} & 38.41 &
        40.88 & 38.87 &
        38.25 & 38.29 & 38.32 \\ \hline
38 & HMM & 21.45 & {\bf 21.46} & 21.60 &
        24.02 & 21.84 &
        21.49 & 21.49 & 21.49 \\ \hline
41 & HMM & 13.91 & {\bf 13.92} & 13.94 &
        14.06 & 14.02 &
        13.98 & 13.98 & 14.02 \\ \hline
44 & HMM & 11.71 & {\bf 11.76} & 12.04 &
        12.62 & 12.70 &
        12.01 & 12.04 & 12.04 \\ \hline
45 & HMM & 24.04 & 24.05 & 24.05 &
        24.05 & {\bf 24.04} &
        {\bf 24.04} & 24.24 & {\bf 24.04} \\ \hline \hline
3 & PNFA & 49.96 & {\bf 50.04} & 50.68 &
        52.27 & 51.35 &
        50.65 & 51.21 & 50.65 \\ \hline
4 & PNFA & 80.82 & {\bf 80.83} &  80.84 &
        82.30 & 80.95 &
       80.93 & 80.89 & 81.02 \\ \hline
8 & PNFA & 81.38 &  {\bf 81.40} & 81.71 &
        91.23 &  {83.01} &
         {84.83} & 82.05 & 82.73 \\ \hline
10 & PNFA & 33.30 & {\bf 33.33} & 34.04 &
        49.51 & 33.65 &
         {35.62} &  {35.04} & 33.47 \\ \hline
12 & PNFA & 21.66 & {\bf 21.66} &  21.77 &
        23.78 & 21.68 &
        21.68 & 22.49 & 21.68 \\ \hline
15 & PNFA & 44.24 & {\bf 44.27} & 44.70 &
        52.29 & 45.10 &
         {48.69} &  {46.80} & 44.66 \\ \hline
17 & PNFA & 47.31 & {\bf 47.35} &  47.92 &
        60.60 & 48.03 &
        47.95 &  {51.13} & 48.11 \\ \hline   
22 & PNFA &25.98 & {\bf 25.99} & 26.08 &
        39.25 & 26.56 &
         {27.26} & 26.61 & 26.37 \\ \hline
29 & PNFA & 24.03 & {\bf 24.04} &  24.11 &
        27.80 & 24.20 &
        24.64 & 24.58 & 24.15 \\ \hline
30 & PNFA & 22.93 & {\bf 22.93} &  23.21 &
        26.05 & 23.47 &
        23.25 & 23.33 & 23.22 \\ \hline
31 & PNFA & 41.21 & {\bf 41.23} & 41.62 &
        43.00 & 42.08 &
        41.51 & 42.27 & 41.60 \\ \hline
34 & PNFA & 19.96 & {\bf 19.97} & 20.54 &
        36.27 &  {25.99} &
         {43.01} &  {26.50} &  {22.63} \\ \hline
37 & PNFA & 20.98 & {\bf 21.00} &  21.02 &
        21.11 & 21.19 &
        21.07 & 21.11 & 21.13 \\ \hline
39 & PNFA & 10.00 & {\bf 10.00} & {\bf 10.00} &
        10.34 & {\bf 10.00} &
        {\bf 10.00} & 10.05 & {\bf 10.00} \\ \hline
43 & PNFA & 32.64 & {\bf 32.72} &  32.78 &
        33.30 & 33.14 &
        32.97 & 32.85 & 33.05 \\ \hline
46 & PNFA & 11.98 & {\bf 11.99} & 12.10 &
        15.55 & 12.50 &
        13.02 & 12.89 & 12.43 \\ \hline

\hline

\end{tabular}
\caption{PAutomaC problems, sorted according to model types (HMM = hidden Markov model, PNFA = probabilistic non-deterministic automaton), and perplexity scores of the solution, best PAutomaC teams, and five FlexFringe evaluation functions.}\label{table}
\end{footnotesize}
\end{table}

\subsection{Alergia improvements}

The results are given in Table~\ref{table}. We first run Alergia as written in the 1994 seminal paper~\cite{carrasco1994learning}. Out of the box Alergia94 performs poorly, and a likely reason for this is the effect of low-frequency counts on the consistency test and the resulting bad merges. Alergia+, including a largest-blue search order, sinks, and our new pooling strategy, is much better in both performance and run-time. We use a \textsc{sink\_count} of 25, which causes FlexFringe to complete the full set of PAutomaC training files in 20 minutes on a single thread at 2.6 GHz. We did not tune the threshold parameter and kept it at its default value of $0.01$. 
We use a \textsc{state\_count} of 15, and a \textsc{symbol\_count} of 10. Note that \textsc{state\_count} has to be lower than \textsc{sink\_count}. If not, states with a frequency below \textsc{state\_count} can be merged with any other state. The merge check will simply return true without performing any test. 
Alergia+ performs particularly well compared to the best-performing state merging approach during the competition (Llorens). The competition winner's Gibbs sampling approach is hard to beat on all problems, particularly those with non-deterministic ground truth models (HMM and PNFA). For those problems, the PNFA learner by team Shibata-Yoshinaka remains the best. Although PDFAs can be used to approximate HMMs, and despite our improvements to state merging, our results confirm that learning a non-deterministic model works better when the data are generated using a non-deterministic model.

For the PDFA ground truth models, the performance of Alergia+ is close to optimal and on par with Llorens and Shibata-Yoshinaka. We emphasize that we did not tune the \textsc{confidence\_bound} parameter used by the statistical tests or run FlexFringe's search procedure to obtain these results. We conclude our improvements to Alergia are effective, resulting in a fast method that obtains good predictive performance scores out of the box when the data originates from a PDFA model. 

\subsection{Other evaluation functions}

We also evaluate the likelihood ratio, MDI, and AIC evaluation functions (consistency and score computations) to demonstrate that the choice of function can greatly affect the obtained performance. In fact, one of the main reasons we developed FlexFringe is to be able to design a new evaluation function quickly. We believe that different problems not only require different parameter settings but often require different evaluation functions. In a way, this is similar to using different loss functions when training neural networks.

We run these different evaluation functions with the same settings for \textsc{sink\_count}, \textsc{symbol\_count}, and \textsc{state\_count}, and their default \textsc{confidence\_bound} parameter. The results from likelihood ratio seem slightly worse than the results we obtained from Alergia+. Although it achieves competitive scores on many problems, the obtained perplexity scores are much larger on several problems. 

MDI also performs worse than Alergia+, though it shows smaller deviations than likelihood ratio, particularly on problem 21.  
Interestingly and unexpectedly, AIC performs best. Ignoring empty lines, the code for AIC is about 20 lines long\footnote{AIC inherits its update routines for counting symbols from Alergia and the log-likelihood and parameter computation from likelihood ratio.}. This result shows the key strength of FlexFringe: the ability to implement new evaluation functions quickly. We did not expect AIC to work so well based on earlier results~\cite{verwer2010efficient}. These results indicate that our pooling and parameter counting strategies positively affect model selection criteria.

\section{Results on HDFS}
\label{sec:results_hdfs}
The HDFS data set~\cite{xu2009detecting} is a well-known data set for evaluating software log anomaly detection algorithms and has, for instance, been used to evaluate the DeepLog anomaly detection framework that is based on neural networks~\cite{du2017deeplog}. We obtained the data from the DeepLog GitHub repository. The training data consists of 4855 (normal) training traces, 16838 abnormal testing traces, and 553366 normal testing traces. We thus see only a small fraction of the normal data at training time. Despite this restriction, DeepLog shows quite good performance on detecting anomalies~\cite{du2017deeplog}: 833 false positives (normal labeled as abnormal) and 619 false negatives (abnormal labeled as normal).

The easiest way to use FlexFringe for anomaly detection is to learn a PDFA for the system behavior under normal circumstances from the training traces. The learned PDFA model $\mathcal{A}$ is then applied to the testing traces $S$ to compute their probabilities $\{\mathcal{A}(s) : s \in S\}$. The final step is to set a decision threshold $t$, the set $\{s : \mathcal{A}(s) \leq t\}$ are then the detected anomalies. In this paper, we set $t$ to $0$, so only traces that at some point try to trigger a transition that does not exist or end in a state without a final probability are labeled as anomalies (``-inf'' as prediction, see Section~\ref{sec:run}).

The first few lines of the training file given to FlexFringe are shown in Figure~\ref{fig:hdfs_data}. As can be seen, this data contains patterns that are quite typical in software systems, such as parallelism (such as the 22 occurring before or after 5 5 5), repetitions (such as 11 9 occurring thrice), and optional sub-processes (such as repetitions of 3s and 4s). The deterministic nature of the models learned by FlexFringe offers advantages over using more complex models such as neural networks. Firstly, automaton models provide insight into the software process that generated the data when visualized. Secondly, learned automata perform competitively  on problems such as sequence prediction and anomaly detection. Thirdly, learning automata is very fast. FlexFringe requires less than a second of training time to return good-performing and insightful models from the HDFS training data. We now present the results of FlexFringe on this data, first in terms of insight and then in terms of performance.

\begin{figure}
    \centering
    \begin{small}\begin{verbatim}
4855 50
1 19 5 5 5 22 11 9 11 9 11 9 26 26 26 23 23 23 21 21 21
1 13 22 5 5 5 11 9 11 9 11 9 26 26 26
1 21 22 5 5 5 26 26 26 11 9 11 9 11 9 2 3 23 23 23 21 21 21
1 13 22 5 5 5 11 9 11 9 11 9 26 26 26
1 31 22 5 5 5 26 26 26 11 9 11 9 11 9 4 3 3 3 4 3 4 3 3 4 3 3 23 23 23 21 21 21
    \end{verbatim}\end{small}
    \caption{The first six lines of the HDFS training data provided to FlexFringe in Abbadingo format~\cite{lang1998results}. The first line gives the number of sequences and the alphabet size. Then, each line presents a sequence by specifying the sequence type, length, and the sequence itself as a list of symbols. All traces have type 1, meaning they are all valid system occurrences. When multiple classes or sequence types are available, one can specify the type here in order to learn a classifier. In this use case, we learn a probabilistic model and do not care about sequence types. From these few sequences, we already see several subprocesses with symbols: 5s-22s at the start, 11s-9s in the middle, and 23s-21s at the end. Optionally, 2s-3s-4s appears before the 23s-21s. FlexFringe will capture such structures and more hidden ones.}
    \label{fig:hdfs_data}
\end{figure}

\begin{figure}
    \centering
    \includegraphics[height=0.84\textheight]{hdfs_train.dat.ff.final_paper_sinks5.pdf}
    \caption{The result of FlexFringe's AIC heuristic using a \textsc{sink\_count} of 5 without tuning on the HDFS data set. There is parallelism at the start (widening at the top), followed by chains of events (left and right of the middle) or more complex repetitions (center middle), and an ending sequence (starting from the state with many incoming transitions). The thicknesses of the states are an indication of their frequency.}
    \label{fig:hdfs_insight_sinks}
\end{figure}

\subsection{Software process insight}
We run the AIC evaluation function out of the box on the training data to get initial insight into the data. The result is shown in Figure~\ref{fig:hdfs_insight_sinks}. We can clearly distinguish subprocesses and parallelism. The top half of the process forms a narrow-wide-narrow shape indicative of parallel executions. The initial parallel processing consisting of 22 and 5 values is followed by three 26s and three pairs of 11s and 9s. These can all be executed in parallel, causing a very wide model containing (at least) one state for every possible set of previously executed values. Around halfway through the model, the processing continues, starting from just two frequent states. Figure~\ref{fig:subplot} shows the subgraph for the initial parallel processing of 22 and 5 values in more detail. A 22 value can and does occur before, during, or after three occurrences of a 5 value. Interestingly, this processing ends in a different state (17 instead of 14) when three 5s follow a 22. Although the subsequent processing is identical, this makes sense due to the large difference in frequency of the subsequent 11 values. The parallel processing of 9s, 11s, and 26s is similar but much more complicated.

The field of process mining~\cite{van2012process} is focused on methods that explicitly model such behavior using Petri Nets. These only require a single transition for a parallel event, such as the 9, 11, and 26 values in the HDFS data. Automata can model parallel behavior, but at a significant cost in model size since every possible set of previously seen values requires a unique state. Since the bias of automaton learning is to minimize this size, it is nice to see that FlexFringe can discover this behavior from only a few thousand traces. In future work, we aim to extend FlexFringe to search for such behavior and complete the obtained models. 
For instance, if we observe $abc$, $cba$, $bac$, and $cab$, we would like to infer that $acb$ is also possible. The current merging routines are unable to do so. Note that although process mining techniques can model parallelism, they have much more problems with modeling sequential context and counting (such as a 5 occurring exactly 3 times).

After the initial two processes (forming the diamond), there are two possible subprocesses: an infrequent long chain of executions 25-18-5-6-26-26-21, which can be repeated, and a frequent process with many repetitions of 2s, 3s, and 4s. These processes can also be skipped, and the repetitions can end at different points. This can be seen by the many transitions to the final process consisting of three optional repetitions of 23s and 21s.

Overall, the learned model provides a lot of insight into the process structure that generated the logs. We could reach similar conclusions simply by looking at the log files, but we cannot look at 4855 log lines in one view; the learned automaton provides such a view. Moreover, it can show patterns that would be hard to find via manual investigation. For instance, after the parallel executions of the 9s, 11s, and 26s, there are two possible futures depending on whether the final symbol is a 9 or 26. In the latter case, starting the 23s and 21s ending sequence is much more likely. When the parallel execution ends with a 9, only 361 out of 1106 traces start this ending. When ending with a 26, these sequences occur 2519 out of 4375 times. This difference causes the learning algorithm to infer there are two states that signify the end of the 9-11-26 parallel execution: state 98 and state 100. These are the frequent (thick-edged) states in the middle left and middle right of the automaton model. Another observation is that this 23-21 ending sequence can be started from many different places in the system, but after the initial parallel executions. This can be seen by the many input transitions to state 102, the frequent state in the bottom middle part of the model.

The model also shows some strange bypasses of this behavior, for instance, the rightmost infrequent path that skips the frequent states after the 9-11-26 parallel executions. This path occurs only twice in the entire training data. Consequently, the statistics used to infer this path are not well estimated. It seems likely that the learning algorithm made an incorrect inference, i.e., these frequent states should not be bypassed. We are currently working on techniques to change FlexFringe's bias and avoid making such mistakes. Note that the only way to identify such issues is by visualizing and reasoning about the obtained models, which is prohibitively hard for many other machine learning models, such as neural nets. A recent line of research aims to overcome this limitation by learning automaton models from complex neural networks~\cite{weiss2018extracting, ayache2019explaining, aalpy}.

\begin{figure}
    \centering
    \includegraphics[height=0.85\textheight]{hdfs_train.dat.ff.final_aic.pdf}
    \caption{The result of FlexFringe's AIC heuristic without sinks and without tuning on the HDFS data set. The model is more convoluted than Figure~\ref{fig:hdfs_insight_sinks} because it also models traces that occur infrequently. Several transitions occur only once in the training data. Although they are a bit more hidden due to such infrequent transitions, the same sub-processes can be found as those visible in Figure~\ref{fig:hdfs_insight_sinks}.}
    \label{fig:hdfs_insight}
\end{figure}

\subsection{Anomaly detection performance}
Out of the box, the AIC model seems to capture the underlying process behavior, and it can be used for anomaly detection. The most straightforward approach, which does not involve setting a decision threshold, is to run the test set through the model and raise an alarm either when a trace ends in a state without any final occurrences or when it tries to trigger a transition that does not exist. This strategy gives 11677 false positives and 8 false negatives. These can be reduced to 4132 false positives and only 1 false negative by learning a PDFA without using sinks. The resulting model (see Figure~\ref{fig:hdfs_insight}) is more convoluted, but the sub-processes found using a \textsc{sink\_count} of 5 are still visible. Using the F1 score as a metric gives a score of 0.89, which is worse than the 0.96 obtained by DeepLog on the same data.

We can, of course, improve this performance by tuning several parameters. Before we do so, it is insightful to understand the cause of the false positives. FlexFringe learns (merges states) by testing whether the future process is independent from the past process. Merging more will generalize more and hence cause fewer false positives. But should this be our aim?

One of the key strengths of learning a deterministic automaton model is that one can easily follow a trace's execution path ~\cite{hammerschmidt2016interpreting}. The simplicity of our anomaly detection setup then allows us to reason whether a detected anomaly is a true or false alarm by traversing the PDFA model or even the merge steps performed by the learning algorithm. This kind of explainable machine learning is very hard to perform using neural networks. Investigating the raised false positives provides us with four frequent types of anomalous traces in the normal test set:

\begin{figure}
    \centering
    \includegraphics[width=\textwidth]{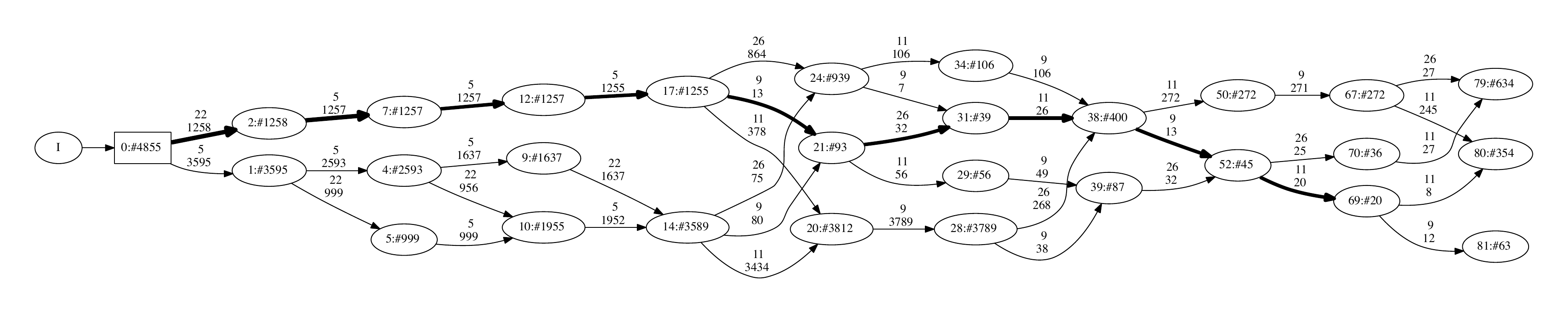}
    \caption{A subplot of the PDFA from Figure~\ref{fig:hdfs_insight} showing paths taken by the mentioned anomalous traces and several parallel states (fin and path counts removed for clarity). State 7 occurs 1257 times in the training data (of 4855 traces), and the subsequent event is always 5 (never 11 as in the different start traces). The infrequent trace reaches state 69 (thick path). There is no outgoing transition with label 26 from state 69. However, we could infer from the parallel paths that the target state should be state 79, i.e., the state reached by swapping the last 11 and 26 events, since it seems these can be executed in parallel. }
    \label{fig:subplot}
\end{figure}

\begin{enumerate}
    \item Not starting with a 22 and three 5s, e.g., 22-5-11-9-5-11-9-5-11-9-26-26-26.
    \item Following an infrequent path, e.g., 22-5-5-5-9-26-11-9-11-26...
    \item Containing symbols not in the train set, e.g., 22-5-5-5-...-3-4-23-23-23-21-21-{\bf20}-21.
    \item Repetition of values, e.g., 5-5-...-4-4-4-3-4-4-4-4-4-4-4-4-4-4-4-4-2-2-...
\end{enumerate}

To facilitate the analysis of these behaviors, we plot a subgraph from Figure~\ref{fig:hdfs_insight} in Figure~\ref{fig:subplot}. The different start traces quickly reach a state without a transition for the next symbol. The listed trace ends after the 22 and 5 symbols. The reached state (nr. 7) occurred 1257 times by traces from the training data, and all of these traces had 5 as their next symbol. We would argue that this is an anomaly that should be raised. In fact, there are only 63 traces that start with 22-5-11 in the entire test and 139761 that start with 22-5. Still, these are counted as false positives when computing the F1 score.

The infrequent paths end in, or traverse, states that occur infrequently. The listed prefix ends after the second 11 symbol in a state that occurs only 20 times and always has a 9 or 11 as the next symbol in the training data. This seems to be no anomaly, and different parameter settings would likely cause a merge of this state and thus possibly provide a transition with label 26. We argue, however, that this is bad practice as learning such an infrequent state is no mistake. Many states are required to model the parallelism in the data, and several of these will be infrequent. Given this parallelism, we actually know what state to target, the one reached by the prefix 22, 5, 5, 5, 9, 26, 11, 9, 26, 11 (state number 79). We swapped the last two symbols. This state occurs much more frequently (634 times), and we could add this transition (from state 69 to 79 with label 26) to the model. In future work, we aim to extend a learned automaton with such 0-occurrence transitions or check for them at test time.

The traces with new symbols are clearly abnormal and should be counted as true negatives rather than false positives. The HDFS data is somewhat strange in that events occur in the test set that never occurred at train time. Also, many of the true positive traces contain such symbols.

Traces with many repetitions do show mistakes made by the learning algorithm. The algorithm has learned that repetitions are possible. A part of the PDFA shows these repetitions as loops, but several possible repetitions are not modeled correctly. Performing more or different merges can change this and potentially remove these false positives. Learning which repetitions are possible and which are not requires more data or a different learning strategy/parameter settings.

The above analysis shows how a PDFA can be used to understand and detect anomalies. Such understanding is useful for investigating alarms, tuning parameters of the learning algorithm, and discovering limitations and potential improvements for the detection pipeline.

\begin{figure}
    \centering
    \includegraphics[width=\textwidth,height=0.8\textheight]{hdfs_train.dat.ff.final_lowthreshold.pdf}
    \caption{The result of FlexFringe's likelihood ratio heuristic with a very low \textsc{confidence\_bound}. It performs many more merges compared to Figure~\ref{fig:hdfs_insight}. As a consequence, it is much harder to interpret. We still see some parallelism at the top, but many self-loops are likely overgeneralizations. Moreover, the many (probably incorrect) long transitions hinder interpretability. In terms of F1-score, it performs better than the model from Figure~\ref{fig:hdfs_insight}, and better than the DeepLog baseline~\cite{du2017deeplog}}
    \label{fig:hdfs_anomaly}
\end{figure}

\subsection{A different learning strategy}

One way to raise fewer false alarms is to perform more merges and thus obtain fewer states that have more outgoing transitions. The AIC evaluation function has no significance parameter (\textsc{confidence\_bound}). Instead, we learn another model using the likelihood ratio evaluation function and a very low \textsc{confidence\_bound} of 1E-15. Other than that, we keep the default settings. The resulting model is displayed in Figure~\ref{fig:hdfs_anomaly}. The model is much less insightful than Figure~\ref{fig:hdfs_insight} and likely overgeneralizes due to all the added loops. It seems to model impossible system behavior, such as infinite loops of 21s. In terms of performance, however, this model achieves 330 false positives and 624 false negatives, i.e., an F1-score of 0.97, outperforming the score achieved by DeepLog.

This demonstrates that automaton learning methods can achieve a performance similar to that of neural network approaches on software log data. We believe the main reason is that software data is highly structured and often deterministic. In the experiments on the PAutomaC data, we also demonstrated that deterministic automata learned using FlexFringe perform very well when the ground truth model is deterministic. Automata are simply good at capturing the type of patterns that occur in deterministic systems.

A key question and challenge for future work is how to treat infrequent states during learning. Is it better to keep them intact to obtain a more interpretable model, or should we merge them and get improved performance at the cost of interpretability? In order to prevent this trade-off, we are currently extending FlexFringe with methods that look for software-specific patterns, such as parallelism and subprocesses. We believe that such extensions will be crucial for obtaining high-performing interpretable models.

\section{Related works}
\label{sec:related}
There exist a lot of different algorithms for learning (P)DFAs. Like FlexFringe, most of these use some form of state consistency based on their future behavior, i.e., a test for a Markov property or Myhill-Nerode congruence. Many algorithms are active. These learn by interacting with a black-box system-under-test (SUT) by providing input and learning from the produced output. Starting from the seminal L* work in~\cite{angluin1987learning} and its successful implementation in the LearnLib tool~\cite{raffelt2005learnlib}, many works have applied and extended this algorithm, e.g., to analyze and reverse engineer protocols~\cite{fiteruau2016combining,fiterau2020analysis} and learn register automata~\cite{isberner2014learning, aarts2015learning}. There exist a few extensions that actively aim to learn probabilistic models, such as PDFAs. A complication is that PDFA learning algorithms rely on the information present in occurrence counts to infer state similarity. In most active learning setups, the algorithm determines these counts through the inputs it provides. An exception is the setup in~\cite{tappler2019based}. They develop a new active learning algorithm that learns a Markov Decision Process from distribution or frequency information outputs. Their algorithm outperforms passive learning from a fixed data set by actively sampling for such information. In reinforcement learning, active and passive learning are actively studied to learn a similar model called a Regular Decision Processes~\cite{brafman2019regular,cipollone2024provably}.

Although active learning is closely related to learning from a data set~\cite{lange2004formal}, since FlexFringe does not yet learn actively, we will not elaborate more on these approaches and refer to~\cite{vaandrager2017model} for an overview of active learning algorithms and their application. Below, we present related algorithms that learn from a data set as input.

\subsection{Algorithms}
We described the main state-merging algorithms FlexFringe builds upon in Section~\ref{sec:FlexFringe}. In the literature, several other approaches exist. A closely related research line consists of different versions of the k-Tails algorithm~\cite{biermann1972synthesis}, essentially a state-merging method that limits consistency until depth k for computational reasons. Moreover, this allows us to infer models from unlabeled data without using probabilities: simply require identical suffixes up to depth k. In the original work, the authors propose to solve this problem using mathematical optimization. Afterward, many greedy versions of this algorithm have been developed and applied to various software logs~\cite{cook1998discovering}. Notable extensions of state merging methods are the declarative specifications~\cite{beschastnikh2013unifying}, learning from concurrent/parallel traces~\cite{beschastnikh2014inferring}, and learning guarded, extended, and timed automata~\cite{mariani2016gk,walkinshaw2016inferring,pastore2017timed, hall2017data}. Several ways to speed up state-merging algorithms have been proposed through divide and conquer and parallel processing~\cite{luo2017inferring, akram2010psma, shin2021prins}. There have also been several proposals to use different search strategies such as beam search~\cite{bugalho2005inference}, genetic algorithms~\cite{lucas2005learning,tsarev2011finite}, satisfiability solving~\cite{heule2013software,zakirzyanov2017finding}, and ant-colony optimization~\cite{chivilikhin2013muacosm}.

Another closely related line of work focuses on spectral learning methods. Spectral learning formulates the PDFA (or weighted automaton) learning problem as finding the spectral decomposition of a Hankel matrix~\cite{balle2014spectral, glaude2015spectral}. Every row in this matrix represents a prefix, every column a suffix, and each cell contains the string probability of the corresponding row and column prefix and suffix. The rows of this matrix correspond to states of the prefix tree. If one row is a multiple of another, the future suffix distribution of the corresponding states is similar, i.e., they can be merged. Instead of searching for such similarities and forcing determinization, spectral methods approximate this using principal component analysis, returning a probabilistic nondeterministic finite state automaton (PNFA). Because non-deterministic models can be in multiple states simultaneously, they are harder to interpret (although typically smaller) than their deterministic counterparts.

Due to their close relationship with hidden Markov models (HMMs)~\cite{dupont2005links}, several approaches exist that infer HMMs instead of PDFAs from trace data. HMMs are typically learned using a form of expectation-maximization known as the  Baum-Welch algorithm~\cite{rabiner1986introduction}. The winner of the PAutomaC challenge by Shibata-Yoshinaka used a related method based on collapsed Gibbs sampling~\cite{blei2006variational, wang2013collapsed} for approximate inference, see~\cite{verwer2014pautomac} for more details. The runner-up, team Llorens, used a state merging method similar to the one described in Section~\ref{sec:FlexFringe}. Special state merging~\cite{stolcke1992hidden} or state splitting~\cite{takami1992successive} algorithms have also been proposed. A notable recent approach~\cite{emam2018inferring} learns accurate probabilistic extended automata using HMMs combined with reinforcement learning.

\subsection{Tools}
There exist several implementations of state-merging algorithms. We list the most popular ones and highlight differences with FlexFringe.

\subsubsection{MINT}~\cite{walkinshaw2016inferring} is a tool for learning extended DFAs. These contain guards on values in addition to symbols. In MINT, these guards are inferred using a classifier from standard machine learning tools, which aims to predict the next event based on features of the current event. When triggering a transition, the guard is used together with the symbol to determine the next state. FlexFringe also contains such functionality, but instead of a classifier, it uses a decision-tree-like construction to determine guards. Moreover, FlexFringe uses the RTI procedure for this construction, which requires consistency for the entire future instead of only the next event. Finally, in MINT, the learning of these guards is performed as preprocessing. In FlexFringe, it is computed on the fly for every blue state (merge candidate). MINT contains several algorithms, including GK-Tails~\cite{lorenzoli2008automatic}, which uses the Daikon invariant inference system~\cite{ernst2007daikon} to learn guards.

\subsubsection{Synoptic and CSight}~\cite{beschastnikh2011synoptic, beschastnikh2014inferring} are tools based on k-Tails style state-merging of non-probabilistic automata. They are focused on learning models for concurrent and distributed systems, contain methods to infer invariants, and can be combined with model checkers to verify these invariants against the learned models. When a model fails to satisfy an invariant, it is updated using counter-example guided abstraction refinement (CEGAR)~\cite{clarke2000counterexample}. Although CEGAR is a common way to implement active learning algorithms, Synoptic and CSight learn from data sets. From the same lab also comes InvariMint~\cite{beschastnikh2013unifying}, a framework for declaratively specifying automaton learning algorithms using properties specified by LTL formulas. Similar specifications in other first-order logic have also been proposed~\cite{bruynooghe2015predicate}. Such specifications are very powerful and allow for much flexibility in designing learning algorithms, as a new algorithm requires just a few lines of code/formulas. Some properties, such as statistical tests, are hard to specify. FlexFringe allows specifications of new evaluation functions by writing code instead of formulas. Currently, FlexFringe does not contain CEGAR-like refinement functionality or methods to mine invariants.

\subsubsection{GI-learning}~\cite{cottone2016gl} is an efficient toolbox for DFA learning algorithms written in C++, including significant speedups due to parallel computation of merge tests. It contains implementations of basic approaches for both active algorithms and algorithms that learn from a data set. It is possible to include more algorithms and different types of automata by extending the classes of these basic approaches. FlexFringe makes this easier by only requiring new consistency check and score method implementations. FlexFringe currently contains no parallel processing methods, but using the union/find data structures (see Section~\ref{sec:FlexFringe}) makes FlexFringe already very efficient.

\subsubsection{LibAlf}~\cite{bollig2010libalf} is a well-known extensive library for automaton learning algorithms, both active and for learning from a data set. It includes many standard and specialized algorithms, for instance, for learning visibly one-counter automata and non-deterministic automata. Like FlexFringe, it is easily extensible but does not include algorithms for learning guards or probabilistic automata.

\subsubsection{AALpy}~\cite{aalpy} is a recent lightweight active automaton learning library written in pure Python. In addition to many active algorithms and optimizations, it contains basic algorithms for learning from a data set. A key feature of AALpy is its ease of use and the many different kinds of models that can be learned, including non-deterministic ones. It is extensible by defining new types of automata and algorithms. It has a different design from FlexFringe in that a new algorithm requires new implementations of the merge routines instead of only the evaluation functions. AALpy currently has no support for inferring transition guards from real-valued input. However, one could run the MINT pre-processing before providing the data to AALpy. FlexFringe contains the RTI+ algorithm~\cite{verwer2010likelihood} that learns guards while running the state merging process.

\subsubsection{LearnLib}~\cite{isberner2015open} is a popular toolkit for active learning of automata, in particular Mealy machines. It has methods to connect to a software system under test by mapping discrete symbols from the automata's alphabet to concrete inputs for the software system, such as network packets. In addition, it contains different model-based testing methods~\cite{lee1996principles} that are used to find counterexamples to a hypothesized automaton and optimized active learning algorithms such as TTT~\cite{isberner2014ttt}. As such, it is frequently used in real-world use cases, see, e.g.,~\cite{fiterau2020analysis}. Extensions for LearnLib exist, such as the ability to learn extended automata~\cite{cassel2015ralib}.

\subsubsection{Sp2Learn}~\cite{arrivailt2017sp2learn} is a library for spectral learning of weighted or probabilistic automata from a data set written in Python. It learns non-deterministic automata, which are typically harder to interpret than deterministic ones but can more efficiently model distributions from non-deterministic systems. Spectral learning can be very effective, as it solves the learning problem using a polynomial-time decomposition algorithm. In contrast, FlexFringe's state-merging methods also run in polynomial time but likely result in a local minimum. Search procedures that aim to find the global optimum are costly to run.

\subsubsection{DISC}~\cite{shvo2020interpretable} is a recent mixed integer linear programming method for learning non-probabilistic automata from a data set. Using mathematical optimization is a promising recent approach for solving machine learning problems such as decision tree learning~\cite{carrizosa2021mathematical, verwer2017learning, bertsimas2017optimal}. FlexFringe contains one such approach based on satisfiability solvers instead of integer programming. An advantage of DISC is that it can handle noisy data due to the use of integer programming, which uses continuous relaxations during its solving procedure. Due to the explicit noise modeling, it can handle some types of non-determinism without requiring additional states. FlexFringe does not explicitly model noise but does allow for more robust evaluation functions, such as impurity metrics used in decision tree learning.

\section{Conclusion}
\label{sec:conclusions}
FlexFringe provides efficient implementations of key state-merging algorithms, including optimizations to get improved results in practice. We presented how to use FlexFringe to learn probabilistic finite state automata (PDFAs) and demonstrated the improvement over a standard implementation of the famous Alergia algorithm on the PAutomaC dataset. 
FlexFringe can learn these using various methods such as EDSM, Alergia, RPNI, MDI, AIC, and different search strategies, and can be used to compare these methods. The kinds of automata and the used evaluation function can be changed by adding a single file to the code base. All that is needed is to specify when a merge is inconsistent and what score to assign to a possible merge. Compared to existing tools, the main restriction is that the learned models have to be deterministic. This is an invariant we use to speed up the state-merging algorithm. Although deterministic models can approximate non-deterministic ones, as our results on the PAutomaC dataset demonstrate, it can be better to learn a non-deterministic model when the data comes from a non-deterministic system.

FlexFringe obtains competitive results on prediction and anomaly tasks thanks to our optimizations. Moreover, the learned models can provide insight into the inner workings of black-box software systems. On trace prediction, our results show FlexFringe performs especially well when the data are generated from a deterministic system. On anomaly detection performance, the model produced by FlexFringe is competitive with an existing method based on neural networks while requiring only seconds of run-time to learn. We believe this is due to software data properties such as little noise and determinism, which favor automaton models.

We demonstrate a clear trade-off between the obtained insight and the (predictive) performance of models. 
Sometimes, it is best to keep data intact, e.g., when there is too little data to determine what learning (merging) step to take. FlexFringe provides techniques such as sinks to prevent the state-merging algorithm from performing incorrect merges, which can be detrimental to obtaining insight, as these often lead to incorrect conclusions. Also, 
merging with little evidence often leads to convoluted models. However, such convoluted models tend to perform better for making predictions due to their increased generalization. This trade-off deserves further study. We expect that there exist better generalization methods for software systems that lead to both improved insight and improved performance.

\newpage

\bibliographystyle{alphaurl}
\bibliography{pdfa_learning}

\newcommand{\etalchar}[1]{$^{#1}$}
\begin{thebibliography}{VTDLH{\etalchar{+}}05}

\bibitem[ABDE17]{arrivailt2017sp2learn}
Denis Arrivault, Dominique Benielli, François Denis, and Rémi Eyraud.
\newblock Sp2learn: A toolbox for the spectral learning of weighted automata.
\newblock In {\em International Conference on Grammatical Inference}, pages
  105--119. PMLR, 2017.

\bibitem[ABDLHE10]{akram2010psma}
Hasan~Ibne Akram, Alban Batard, Colin De~La~Higuera, and Claudia Eckert.
\newblock {PSMA}: A parallel algorithm for learning regular languages.
\newblock In {\em NIPS workshop on learning on cores, clusters and clouds},
  2010.

\bibitem[ABL02]{ammons2002mining}
Glenn Ammons, Rastislav Bodik, and James~R Larus.
\newblock Mining specifications.
\newblock {\em ACM Sigplan Notices}, 37(1):4--16, 2002.

\bibitem[ACD91]{alur1991model}
Rajeev Alur, Costas Courcoubetis, and David Dill.
\newblock Model-checking for probabilistic real-time systems.
\newblock In {\em Automata, Languages and Programming: 18th International
  Colloquium Madrid, Spain, July 8--12, 1991 Proceedings 18}, pages 115--126.
  Springer, 1991.

\bibitem[ACS04]{abela2004mutually}
John Abela, Fran{\c{c}}ois Coste, and Sandro Spina.
\newblock Mutually compatible and incompatible merges for the search of the
  smallest consistent {DFA}.
\newblock In {\em International Colloquium on Grammatical Inference}, pages
  28--39. Springer, 2004.

\bibitem[AEG19]{ayache2019explaining}
St{\'e}phane Ayache, R{\'e}mi Eyraud, and No{\'e} Goudian.
\newblock Explaining black boxes on sequential data using weighted automata.
\newblock In {\em International Conference on Grammatical Inference}, pages
  81--103. PMLR, 2019.

\bibitem[AFBKV15]{aarts2015learning}
Fides Aarts, Paul Fiterau-Brostean, Harco Kuppens, and Frits Vaandrager.
\newblock Learning register automata with fresh value generation.
\newblock In {\em International Colloquium on Theoretical Aspects of
  Computing}, pages 165--183. Springer, 2015.

\bibitem[AJ06]{adriaans2006using}
Pieter Adriaans and Ceriel Jacobs.
\newblock Using {MDL} for grammar induction.
\newblock In {\em Grammatical Inference: Algorithms and Applications: 8th
  International Colloquium, ICGI}, pages 293--306. Springer, 2006.

\bibitem[Aka74]{akaike1974new}
Hirotugu Akaike.
\newblock A new look at the statistical model identification.
\newblock {\em IEEE Transactions on Automatic Control}, 19(6):716--723, 1974.

\bibitem[Ang87]{angluin1987learning}
Dana Angluin.
\newblock Learning regular sets from queries and counterexamples.
\newblock {\em Information and Computation}, 75(2):87--106, 1987.

\bibitem[ANV11]{antunes2011reverse}
Joao Antunes, Nuno Neves, and Paulo Verissimo.
\newblock Reverse engineering of protocols from network traces.
\newblock In {\em 2011 18th Working Conference on Reverse Engineering}, pages
  169--178. IEEE, 2011.

\bibitem[AV07]{adriaans2007power}
Pieter Adriaans and Paul Vitanyi.
\newblock The power and perils of {MDL}.
\newblock In {\em 2007 IEEE International Symposium on Information Theory},
  pages 2216--2220. IEEE, 2007.

\bibitem[AV10]{aarts2010learning}
Fides Aarts and Frits Vaandrager.
\newblock Learning i/o automata.
\newblock In {\em International Conference on Concurrency Theory}, pages
  71--85. Springer, 2010.

\bibitem[BABE11]{beschastnikh2011synoptic}
Ivan Beschastnikh, Jenny Abrahamson, Yuriy Brun, and Michael~D Ernst.
\newblock Synoptic: Studying logged behavior with inferred models.
\newblock In {\em Proceedings of the 19th ACM SIGSOFT Symposium and the 13th
  European Conference on Foundations of Software Engineering}, pages 448--451,
  2011.

\bibitem[BBA{\etalchar{+}}13]{beschastnikh2013unifying}
Ivan Beschastnikh, Yuriy Brun, Jenny Abrahamson, Michael~D Ernst, and Arvind
  Krishnamurthy.
\newblock Unifying {FSM}-inference algorithms through declarative
  specification.
\newblock In {\em 2013 35th International Conference on Software Engineering
  (ICSE)}, pages 252--261. IEEE, 2013.

\bibitem[BBB{\etalchar{+}}15]{bruynooghe2015predicate}
Maurice Bruynooghe, Hendrik Blockeel, Bart Bogaerts, Broes De~Cat, Stef
  De~Pooter, Joachim Jansen, Anthony Labarre, Jan Ramon, Marc Denecker, and
  Sicco Verwer.
\newblock Predicate logic as a modeling language: modeling and solving some
  machine learning and data mining problems with {IDP3}.
\newblock {\em Theory and Practice of Logic Programming}, 15(6):783--817, 2015.

\bibitem[BBEK14]{beschastnikh2014inferring}
Ivan Beschastnikh, Yuriy Brun, Michael~D Ernst, and Arvind Krishnamurthy.
\newblock Inferring models of concurrent systems from logs of their behavior
  with {CSight}.
\newblock In {\em Proceedings of the 36th International Conference on Software
  Engineering}, pages 468--479, 2014.

\bibitem[BCG13]{balle2013learning}
Borja Balle, Jorge Castro, and Ricard Gavald{\`a}.
\newblock Learning probabilistic automata: A study in state distinguishability.
\newblock {\em Theoretical Computer Science}, 473:46--60, 2013.

\bibitem[BCLQ14]{balle2014spectral}
Borja Balle, Xavier Carreras, Franco~M. Luque, and Ariadna Quattoni.
\newblock Spectral learning of weighted automata.
\newblock {\em Machine Learning}, 96(1):33--63, 2014.

\bibitem[BD17]{bertsimas2017optimal}
Dimitris Bertsimas and Jack Dunn.
\newblock Optimal classification trees.
\newblock {\em Machine Learning}, 106(7):1039--1082, 2017.

\bibitem[BDG{\etalchar{+}}19]{brafman2019regular}
Ronen~I Brafman, Giuseppe De~Giacomo, et~al.
\newblock Regular decision processes: A model for non-{M}arkovian domains.
\newblock In {\em IJCAI}, pages 5516--5522, 2019.

\bibitem[BF72]{biermann1972synthesis}
Alan~W Biermann and Jerome~A Feldman.
\newblock On the synthesis of finite-state machines from samples of their
  behavior.
\newblock {\em IEEE Transactions on Computers}, 100(6):592--597, 1972.

\bibitem[BIPT09]{bertolino2009automatic}
Antonia Bertolino, Paola Inverardi, Patrizio Pelliccione, and Massimo Tivoli.
\newblock Automatic synthesis of behavior protocols for composable
  web-services.
\newblock In {\em Proceedings of the 7th Joint Meeting of the European Software
  Engineering Conference and the ACM SIGSOFT Symposium on the Foundations of
  Software Engineering}, pages 141--150, 2009.

\bibitem[BJ06]{blei2006variational}
David~M Blei and Michael~I Jordan.
\newblock Variational inference for {D}irichlet process mixtures.
\newblock 2006.

\bibitem[BKK{\etalchar{+}}10]{bollig2010libalf}
Benedikt Bollig, Joost-Pieter Katoen, Carsten Kern, Martin Leucker, Daniel
  Neider, and David~R Piegdon.
\newblock libalf: The automata learning framework.
\newblock In {\em International Conference on Computer Aided Verification},
  pages 360--364. Springer, 2010.
\newblock URL: \url{https://github.com/libalf/libalf}.

\bibitem[BO05]{bugalho2005inference}
Miguel Bugalho and Arlindo~L Oliveira.
\newblock Inference of regular languages using state merging algorithms with
  search.
\newblock {\em Pattern Recognition}, 38(9):1457--1467, 2005.

\bibitem[CBP{\etalchar{+}}11]{cho2011mace}
Chia~Yuan Cho, Domagoj Babi{\'c}, Pongsin Poosankam, Kevin~Zhijie Chen,
  Edward~XueJun Wu, and Dawn Song.
\newblock {MACE}:model-inference-assisted concolic exploration for protocol and
  vulnerability discovery.
\newblock In {\em 20th USENIX Security Symposium (USENIX Security 11)}, 2011.

\bibitem[CBVR22]{cao2022learning}
Clinton Cao, Agathe Blaise, Sicco Verwer, and Filippo Rebecchi.
\newblock Learning state machines to monitor and detect anomalies on a
  {K}ubernetes cluster.
\newblock In {\em Proceedings of the 17th International Conference on
  Availability, Reliability and Security}, pages 1--9, 2022.

\bibitem[CG08]{castro2008towards}
Jorge Castro and Ricard Gavalda.
\newblock Towards feasible {PAC}-learning of probabilistic deterministic finite
  automata.
\newblock In {\em International Colloquium on Grammatical Inference}, pages
  163--174. Springer, 2008.

\bibitem[CGJ{\etalchar{+}}00]{clarke2000counterexample}
Edmund Clarke, Orna Grumberg, Somesh Jha, Yuan Lu, and Helmut Veith.
\newblock Counterexample-guided abstraction refinement.
\newblock In {\em International Conference on Computer Aided Verification},
  pages 154--169. Springer, 2000.

\bibitem[CHJ15]{cassel2015ralib}
Sofia Cassel, Falk Howar, and Bengt Jonsson.
\newblock {RALib}: A learnlib extension for inferring efsms.
\newblock {\em DIFTS}, 5, 2015.

\bibitem[CJRT24]{cipollone2024provably}
Roberto Cipollone, Anders Jonsson, Alessandro Ronca, and Mohammad~Sadegh
  Talebi.
\newblock Provably efficient offline reinforcement learning in regular decision
  processes.
\newblock {\em Advances in Neural Information Processing Systems}, 36, 2024.

\bibitem[CKW07]{cui2007discoverer}
Weidong Cui, Jayanthkumar Kannan, and Helen~J Wang.
\newblock Discoverer: Automatic protocol reverse engineering from network
  traces.
\newblock In {\em USENIX Security Symposium}, pages 1--14, 2007.

\bibitem[CMRRM21]{carrizosa2021mathematical}
Emilio Carrizosa, Cristina Molero-R{\'\i}o, and Dolores Romero~Morales.
\newblock Mathematical optimization in classification and regression trees.
\newblock {\em Transactions in Operations Research (TOP)}, 29(1):5--33, 2021.

\bibitem[CO94]{carrasco1994learning}
Rafael~C Carrasco and Jose Oncina.
\newblock Learning stochastic regular grammars by means of a state merging
  method.
\newblock In {\em International Colloquium on Grammatical Inference}, pages
  139--152. Springer, 1994.

\bibitem[COP16]{cottone2016gl}
Pietro Cottone, Marco Ortolani, and Gabriele Pergola.
\newblock Gl-learning: an optimized framework for grammatical inference.
\newblock In {\em Proceedings of the 17th International Conference on Computer
  Systems and Technologies 2016}, pages 339--346, 2016.
\newblock URL: \url{https://github.com/piecot/GI-learning}.

\bibitem[CT04]{clark2004pac}
Alexander Clark and Franck Thollard.
\newblock {PAC}-learnability of probabilistic deterministic finite state
  automata.
\newblock {\em Journal of Machine Learning Research}, 5(May):473--497, 2004.

\bibitem[CU13]{chivilikhin2013muacosm}
Daniil Chivilikhin and Vladimir Ulyantsev.
\newblock {MuACOsm}: a new mutation-based ant colony optimization algorithm for
  learning finite-state machines.
\newblock In {\em Proceedings of the 15th Annual Conference on Genetic and
  Evolutionary Computation}, pages 511--518, 2013.

\bibitem[CW98]{cook1998discovering}
Jonathan~E Cook and Alexander~L Wolf.
\newblock Discovering models of software processes from event-based data.
\newblock {\em ACM Transactions on Software Engineering and Methodology
  (TOSEM)}, 7(3):215--249, 1998.

\bibitem[CWKK09]{comparetti2009prospex}
Paolo~Milani Comparetti, Gilbert Wondracek, Christopher Kruegel, and Engin
  Kirda.
\newblock Prospex: Protocol specification extraction.
\newblock In {\em 2009 30th IEEE Symposium on Security and Privacy}, pages
  110--125. IEEE, 2009.

\bibitem[DDE05]{dupont2005links}
Pierre Dupont, Fran{\c{c}}ois Denis, and Yann Esposito.
\newblock Links between probabilistic automata and hidden {M}arkov models:
  probability distributions, learning models and induction algorithms.
\newblock {\em Pattern Recognition}, 38(9):1349--1371, 2005.

\bibitem[DlH10]{higuera2010book}
Colin De~la Higuera.
\newblock {\em Grammatical inference: learning automata and grammars}.
\newblock Cambridge University Press, 2010.

\bibitem[DLZS17]{du2017deeplog}
Min Du, Feifei Li, Guineng Zheng, and Vivek Srikumar.
\newblock Deeplog: Anomaly detection and diagnosis from system logs through
  deep learning.
\newblock In {\em Proceedings of the 2017 ACM SIGSAC Conference on Computer and
  Communications Security}, pages 1285--1298, 2017.

\bibitem[DRP15]{de2015protocol}
Joeri De~Ruiter and Erik Poll.
\newblock Protocol state fuzzing of {TLS} implementations.
\newblock In {\em 24th USENIX Security Symposium (USENIX Security 15)}, pages
  193--206, 2015.

\bibitem[EM18]{emam2018inferring}
Seyedeh~Sepideh Emam and James Miller.
\newblock Inferring extended probabilistic finite-state automaton models from
  software executions.
\newblock {\em ACM Transactions on Software Engineering and Methodology
  (TOSEM)}, 27(1):1--39, 2018.

\bibitem[EPG{\etalchar{+}}07]{ernst2007daikon}
Michael~D Ernst, Jeff~H Perkins, Philip~J Guo, Stephen McCamant, Carlos
  Pacheco, Matthew~S Tschantz, and Chen Xiao.
\newblock The {D}aikon system for dynamic detection of likely invariants.
\newblock {\em Science of Computer Programming}, 69(1-3):35--45, 2007.

\bibitem[FBJM{\etalchar{+}}20]{fiterau2020analysis}
Paul Fiterau-Brostean, Bengt Jonsson, Robert Merget, Joeri De~Ruiter,
  Konstantinos Sagonas, and Juraj Somorovsky.
\newblock Analysis of {DTLS} implementations using protocol state fuzzing.
\newblock In {\em 29th USENIX Security Symposium (USENIX Security 20)}, pages
  2523--2540, 2020.

\bibitem[FBJV16]{fiteruau2016combining}
Paul Fiter{\u{a}}u-Bro{\c{s}}tean, Ramon Janssen, and Frits Vaandrager.
\newblock Combining model learning and model checking to analyze {TCP}
  implementations.
\newblock In {\em International Conference on Computer Aided Verification},
  pages 454--471. Springer, 2016.

\bibitem[FBLP{\etalchar{+}}17]{fiteruau2017model}
Paul Fiter{\u{a}}u-Bro{\c{s}}tean, Toon Lenaerts, Erik Poll, Joeri de~Ruiter,
  Frits Vaandrager, and Patrick Verleg.
\newblock Model learning and model checking of {SSH} implementations.
\newblock In {\em Proceedings of the 24th ACM SIGSOFT International SPIN
  Symposium on Model Checking of Software}, pages 142--151, 2017.

\bibitem[GEP15]{glaude2015spectral}
Hadrien Glaude, Cyrille Enderli, and Olivier Pietquin.
\newblock Spectral learning with proper probabilities for finite state
  automation.
\newblock In {\em ASRU 2015-Automatic Speech Recognition and Understanding
  Workshop}. IEEE, 2015.

\bibitem[Gol78]{gold1978complexity}
E~Mark Gold.
\newblock Complexity of automaton identification from given data.
\newblock {\em Information and Control}, 37(3):302--320, 1978.

\bibitem[HMS{\etalchar{+}}16]{hammerschmidt2016efficient}
Christian Hammerschmidt, Samuel Marchal, Radu State, Gaetano Pellegrino, and
  Sicco Verwer.
\newblock Efficient learning of communication profiles from {IP} flow records.
\newblock In {\em 2016 IEEE 41st Conference on Local Computer Networks (LCN)},
  pages 559--562. IEEE, 2016.

\bibitem[HMU01]{hopcroft2001introduction}
John~E Hopcroft, Rajeev Motwani, and Jeffrey~D Ullman.
\newblock Introduction to automata theory, languages, and computation.
\newblock {\em ACM Sigact News}, 32(1):60--65, 2001.

\bibitem[HV10]{heule2010exact}
Marijn~JH Heule and Sicco Verwer.
\newblock Exact {DFA} identification using {SAT} solvers.
\newblock In {\em International Colloquium on Grammatical Inference}, pages
  66--79. Springer, 2010.

\bibitem[HV13]{heule2013software}
Marijn~JH Heule and Sicco Verwer.
\newblock Software model synthesis using satisfiability solvers.
\newblock {\em Empirical Software Engineering}, 18(4):825--856, 2013.

\bibitem[HVLS16]{hammerschmidt2016interpreting}
Christian~Albert Hammerschmidt, Sicco Verwer, Qin Lin, and Radu State.
\newblock Interpreting finite automata for sequential data.
\newblock {\em arXiv preprint arXiv:1611.07100}, 2016.

\bibitem[HW16]{hall2017data}
Mathew Hall and Neil Walkinshaw.
\newblock Data and analysis code for {GP} {EFSM} inference.
\newblock In {\em IEEE International Conference on Software Maintenance and
  Evolution (ICSME)}, pages 611--611. IEEE, 2016.

\bibitem[IHS14a]{isberner2014learning}
Malte Isberner, Falk Howar, and Bernhard Steffen.
\newblock Learning register automata: from languages to program structures.
\newblock {\em Machine Learning}, 96(1):65--98, 2014.

\bibitem[IHS14b]{isberner2014ttt}
Malte Isberner, Falk Howar, and Bernhard Steffen.
\newblock The {TTT} algorithm: a redundancy-free approach to active automata
  learning.
\newblock In {\em International Conference on Runtime Verification}, pages
  307--322. Springer, 2014.

\bibitem[IHS15]{isberner2015open}
Malte Isberner, Falk Howar, and Bernhard Steffen.
\newblock The open-source {LearnLib}.
\newblock In {\em International Conference on Computer Aided Verification},
  pages 487--495. Springer, 2015.

\bibitem[ISBF07]{ingham2007learning}
Kenneth~L Ingham, Anil Somayaji, John Burge, and Stephanie Forrest.
\newblock Learning {DFA} representations of {HTTP} for protecting web
  applications.
\newblock {\em Computer Networks}, 51(5):1239--1255, 2007.

\bibitem[KABP14]{klerx2014model}
Timo Klerx, Maik Anderka, Hans~Kleine B{\"u}ning, and Steffen Priesterjahn.
\newblock Model-based anomaly detection for discrete event systems.
\newblock In {\em 2014 IEEE 26th International Conference on Tools with
  Artificial Intelligence}, pages 665--672. IEEE, 2014.

\bibitem[Lan99]{lang1999faster}
Kevin~J Lang.
\newblock Faster algorithms for finding minimal consistent dfas.
\newblock {\em NEC Research Institute, Tech. Rep}, 1999.

\bibitem[LAVM18]{lin2018tabor}
Qin Lin, Sridha Adepu, Sicco Verwer, and Aditya Mathur.
\newblock {TABOR}: A graphical model-based approach for anomaly detection in
  industrial control systems.
\newblock In {\em Proceedings of the 2018 on Asia Conference on Computer and
  Communications Security}, pages 525--536, 2018.

\bibitem[LHG17]{luo2017inferring}
Chen Luo, Fei He, and Carlo Ghezzi.
\newblock Inferring software behavioral models with {MapReduce}.
\newblock {\em Science of Computer Programming}, 145:13--36, 2017.

\bibitem[LHPV16]{lin2016short}
Qin Lin, Christian Hammerschmidt, Gaetano Pellegrino, and Sicco Verwer.
\newblock Short-term time series forecasting with regression automata.
\newblock 2016.

\bibitem[LMP08]{lorenzoli2008automatic}
Davide Lorenzoli, Leonardo Mariani, and Mauro Pezz{\`e}.
\newblock Automatic generation of software behavioral models.
\newblock In {\em Proceedings of the 30th International Conference on Software
  Engineering}, pages 501--510, 2008.

\bibitem[LPP98]{lang1998results}
Kevin~J Lang, Barak~A Pearlmutter, and Rodney~A Price.
\newblock Results of the {A}bbadingo one {DFA} learning competition and a new
  evidence-driven state merging algorithm.
\newblock In {\em International Colloquium on Grammatical Inference}, pages
  1--12. Springer, 1998.

\bibitem[LR05]{lucas2005learning}
Simon~M Lucas and T~Jeff Reynolds.
\newblock Learning deterministic finite automata with a smart state labeling
  evolutionary algorithm.
\newblock {\em IEEE Transactions on Pattern Analysis and Machine Intelligence},
  27(7):1063--1074, 2005.

\bibitem[LVD20]{lin2020safety}
Qin Lin, Sicco Verwer, and John Dolan.
\newblock Safety verification of a data-driven adaptive cruise controller.
\newblock In {\em 2020 IEEE Intelligent Vehicles Symposium (IV)}, pages
  2146--2151. IEEE, 2020.

\bibitem[LY96]{lee1996principles}
David Lee and Mihalis Yannakakis.
\newblock Principles and methods of testing finite state machines-a survey.
\newblock {\em Proceedings of the IEEE}, 84(8):1090--1123, 1996.

\bibitem[LZ04]{lange2004formal}
Steffen Lange and Sandra Zilles.
\newblock Formal language identification: Query learning vs. {G}old-style
  learning.
\newblock {\em Information Processing Letters}, 91(6):285--292, 2004.

\bibitem[Mai14]{maier2014online}
Alexander Maier.
\newblock Online passive learning of timed automata for cyber-physical
  production systems.
\newblock In {\em 2014 12th IEEE International Conference on Industrial
  Informatics (INDIN)}, pages 60--66. IEEE, 2014.

\bibitem[MAP{\etalchar{+}}21]{aalpy}
Edi Mu\v{s}kardin, Bernhard~K. Aichernig, Ingo Pill, Andrea Pferscher, and
  Martin Tappler.
\newblock {AALpy}: An active automata learning library.
\newblock In {\em Automated Technology for Verification and Analysis - 19th
  International Symposium, {ATVA} 2021, Gold Coast, Australia, October 18-22,
  2021, Proceedings}, Lecture Notes in Computer Science. Springer, 2021.
\newblock URL: \url{https://github.com/DES-Lab/AALpy}.

\bibitem[MPS16]{mariani2016gk}
Leonardo Mariani, Mauro Pezz{\`e}, and Mauro Santoro.
\newblock Gk-tail+ an efficient approach to learn software models.
\newblock {\em IEEE Transactions on Software Engineering}, 43(8):715--738,
  2016.

\bibitem[NN98]{norris1998markov}
James~R Norris and James~Robert Norris.
\newblock {\em {M}arkov chains}.
\newblock Number~2. Cambridge University Press, 1998.

\bibitem[NSV{\etalchar{+}}12]{niggemann2012learning}
Oliver Niggemann, Benno Stein, Asmir Vodencarevic, Alexander Maier, and
  Hans~Kleine B{\"u}ning.
\newblock Learning behavior models for hybrid timed systems.
\newblock In {\em Twenty-Sixth AAAI Conference on Artificial Intelligence},
  2012.

\bibitem[NVMY21]{nadeem2021alert}
Azqa Nadeem, Sicco Verwer, Stephen Moskal, and Shanchieh~Jay Yang.
\newblock Alert-driven attack graph generation using {S-PDFA}.
\newblock {\em IEEE Transactions on Dependable and Secure Computing}, 2021.

\bibitem[OG92]{oncina1992inferring}
Jos{\'e} Oncina and Pedro Garcia.
\newblock Inferring regular languages in polynomial updated time.
\newblock In {\em Pattern recognition and image analysis: selected papers from
  the IVth Spanish Symposium}, pages 49--61. World Scientific, 1992.

\bibitem[OS98]{oliveira1998efficient}
Arlindo~L Oliveira and Joao P~Marques Silva.
\newblock Efficient search techniques for the inference of minimum size finite
  automata.
\newblock In {\em Proceedings. String Processing and Information Retrieval: A
  South American Symposium (Cat. No. 98EX207)}, pages 81--89. IEEE, 1998.

\bibitem[PLHV17]{pellegrino2017learning}
Gaetano Pellegrino, Qin Lin, Christian Hammerschmidt, and Sicco Verwer.
\newblock Learning behavioral fingerprints from netflows using timed automata.
\newblock In {\em 2017 IFIP/IEEE Symposium on Integrated Network and Service
  Management (IM)}, pages 308--316. IEEE, 2017.

\bibitem[PMM17]{pastore2017timed}
Fabrizio Pastore, Daniela Micucci, and Leonardo Mariani.
\newblock Timed k-tail: Automatic inference of timed automata.
\newblock In {\em 2017 IEEE International Conference on Software Testing,
  Verification and Validation (ICST)}, pages 401--411. IEEE, 2017.

\bibitem[PW93]{pitt1993minimum}
Leonard Pitt and Manfred~K Warmuth.
\newblock The minimum consistent {DFA} problem cannot be approximated within
  any polynomial.
\newblock {\em Journal of the ACM (JACM)}, 40(1):95--142, 1993.

\bibitem[RJ86]{rabiner1986introduction}
Lawrence Rabiner and Biinghwang Juang.
\newblock An introduction to hidden {M}arkov models.
\newblock {\em IEEE ASSP Magazine}, 3(1):4--16, 1986.

\bibitem[RSB05]{raffelt2005learnlib}
Harald Raffelt, Bernhard Steffen, and Therese Berg.
\newblock {LearnLib}: A library for automata learning and experimentation.
\newblock In {\em Proceedings of the 10th International Workshop on Formal
  Methods for Industrial Critical Systems}, pages 62--71, 2005.

\bibitem[SBB21]{shin2021prins}
Donghwan Shin, Domenico Bianculli, and Lionel Briand.
\newblock {PRINS}: Scalable model inference for component-based system logs.
\newblock {\em arXiv preprint arXiv:2106.01987}, 2021.

\bibitem[SG09]{shahbaz2009inferring}
Muzammil Shahbaz and Roland Groz.
\newblock Inferring {M}ealy machines.
\newblock In {\em International Symposium on Formal Methods}, pages 207--222.
  Springer, 2009.

\bibitem[SK14]{schmidt2014online}
Jana Schmidt and Stefan Kramer.
\newblock Online induction of probabilistic real-time automata.
\newblock {\em Journal of Computer Science and Technology}, 29(3):345--360,
  2014.

\bibitem[SLTIM20]{shvo2020interpretable}
Maayan Shvo, Andrew~C Li, Rodrigo Toro~Icarte, and Sheila~A McIlraith.
\newblock Interpretable sequence classification via discrete optimization.
\newblock {\em arXiv preprint arXiv:2010.02819}, 2020.

\bibitem[SO92]{stolcke1992hidden}
Andreas Stolcke and Stephen Omohundro.
\newblock Hidden {M}arkov model induction by {B}ayesian model merging.
\newblock {\em Advances in Neural Information Processing systems}, 5, 1992.

\bibitem[TAB{\etalchar{+}}19]{tappler2019based}
Martin Tappler, Bernhard~K Aichernig, Giovanni Bacci, Maria Eichlseder, and
  Kim~G Larsen.
\newblock L*-based learning of markov decision processes.
\newblock In {\em International Symposium on Formal Methods}, pages 651--669.
  Springer, 2019.

\bibitem[TE11]{tsarev2011finite}
Fedor Tsarev and Kirill Egorov.
\newblock Finite state machine induction using genetic algorithm based on
  testing and model checking.
\newblock In {\em Proceedings of the 13th Annual Conference Companion on
  Genetic and Evolutionary Computation}, pages 759--762, 2011.

\bibitem[Tho00]{thollard2000probabilistic}
Franck~Thollard Thollard.
\newblock Probabilistic {DFA} inference using kullback-leibler divergence and
  minimality.
\newblock In {\em In Seventeenth International Conference on Machine Learning}.
  Citeseer, 2000.

\bibitem[TS92]{takami1992successive}
Jun-ichi Takami and Shigeki Sagayama.
\newblock A successive state splitting algorithm for efficient allophone
  modeling.
\newblock In {\em Acoustics, Speech, and Signal Processing, IEEE International
  Conference on}, volume~1, pages 573--576. IEEE Computer Society, 1992.

\bibitem[Vaa17]{vaandrager2017model}
Frits Vaandrager.
\newblock Model learning.
\newblock {\em Communications of the ACM}, 60(2):86--95, 2017.

\bibitem[VDA12]{van2012process}
Wil Van Der~Aalst.
\newblock Process mining: Overview and opportunities.
\newblock {\em ACM Transactions on Management Information Systems (TMIS)},
  3(2):1--17, 2012.

\bibitem[VEDLH14]{verwer2014pautomac}
Sicco Verwer, R{\'e}mi Eyraud, and Colin De~La~Higuera.
\newblock Pautomac: a probabilistic automata and hidden markov models learning
  competition.
\newblock {\em Machine learning}, 96(1):129--154, 2014.

\bibitem[Ver10]{verwer2010efficient}
Sicco Verwer.
\newblock Efficient identification of timed automata: Theory and practice.
\newblock 2010.

\bibitem[VH17]{verwer2017flexfringe}
Sicco Verwer and Christian~A Hammerschmidt.
\newblock {FlexFringe}: a passive automaton learning package.
\newblock In {\em 2017 IEEE International Conference on Software Maintenance
  and Evolution (ICSME)}, pages 638--642. IEEE, 2017.

\bibitem[VTDLH{\etalchar{+}}05]{vidal2005probabilistic}
Enrique Vidal, Franck Thollard, Colin De~La~Higuera, Francisco Casacuberta, and
  Rafael~C Carrasco.
\newblock Probabilistic finite-state machines-part i.
\newblock {\em IEEE Transactions on Pattern Analysis and Machine Intelligence},
  27(7):1013--1025, 2005.

\bibitem[VWW10]{verwer2010likelihood}
Sicco Verwer, Mathijs~de Weerdt, and Cees Witteveen.
\newblock A likelihood-ratio test for identifying probabilistic deterministic
  real-time automata from positive data.
\newblock In {\em International Colloquium on Grammatical Inference}, pages
  203--216. Springer, 2010.

\bibitem[VZ17]{verwer2017learning}
Sicco Verwer and Yingqian Zhang.
\newblock Learning decision trees with flexible constraints and objectives
  using integer optimization.
\newblock In {\em International Conference on AI and OR Techniques in
  Constraint Programming for Combinatorial Optimization Problems}, pages
  94--103. Springer, 2017.

\bibitem[WB13]{wang2013collapsed}
Pengyu Wang and Phil Blunsom.
\newblock Collapsed variational {B}ayesian inference for hidden {M}arkov
  models.
\newblock In {\em Artificial Intelligence and Statistics}, pages 599--607.
  PMLR, 2013.

\bibitem[WGY18]{weiss2018extracting}
Gail Weiss, Yoav Goldberg, and Eran Yahav.
\newblock Extracting automata from recurrent neural networks using queries and
  counterexamples.
\newblock In {\em International Conference on Machine Learning}, pages
  5247--5256. PMLR, 2018.

\bibitem[WLD{\etalchar{+}}13]{walkinshaw2013stamina}
Neil Walkinshaw, Bernard Lambeau, Christophe Damas, Kirill Bogdanov, and Pierre
  Dupont.
\newblock {STAMINA}: a competition to encourage the development and assessment
  of software model inference techniques.
\newblock {\em Empirical Software Engineering}, 18(4):791--824, 2013.

\bibitem[WRSL21]{watanabe2021probabilistic}
Kandai Watanabe, Nicholas Renninger, Sriram Sankaranarayanan, and Morteza
  Lahijanian.
\newblock Probabilistic specification learning for planning with safety
  constraints.
\newblock In {\em 2021 IEEE/RSJ International Conference on Intelligent Robots
  and Systems (IROS)}, pages 6558--6565. IEEE, 2021.

\bibitem[WTD16]{walkinshaw2016inferring}
Neil Walkinshaw, Ramsay Taylor, and John Derrick.
\newblock Inferring extended finite state machine models from software
  executions.
\newblock {\em Empirical Software Engineering}, 21(3):811--853, 2016.
\newblock URL: \url{https://github.com/neilwalkinshaw/mintframework}.

\bibitem[XHF{\etalchar{+}}09]{xu2009detecting}
Wei Xu, Ling Huang, Armando Fox, David Patterson, and Michael~I Jordan.
\newblock Detecting large-scale system problems by mining console logs.
\newblock In {\em Proceedings of the ACM SIGOPS 22nd Symposium on Operating
  Systems Principles}, pages 117--132, 2009.

\bibitem[YAS{\etalchar{+}}19]{yang2019improving}
Nan Yang, Kousar Aslam, Ramon Schiffelers, Leonard Lensink, Dennis Hendriks,
  Loek Cleophas, and Alexander Serebrenik.
\newblock Improving model inference in industry by combining active and passive
  learning.
\newblock In {\em 2019 IEEE 26th International Conference on Software Analysis,
  Evolution and Reengineering (SANER)}, pages 253--263. IEEE, 2019.

\bibitem[ZSU17]{zakirzyanov2017finding}
Ilya Zakirzyanov, Anatoly Shalyto, and Vladimir Ulyantsev.
\newblock Finding all minimum-size {DFA} consistent with given examples:
  {SAT}-based approach.
\newblock In {\em International Conference on Software Engineering and Formal
  Methods}, pages 117--131. Springer, 2017.

\end{thebibliography}

\end{document}